\documentclass[lettersize,journal]{IEEEtran}

\usepackage{amsmath}  
\usepackage{amsfonts} 
\usepackage{mathtools} 
\usepackage{graphicx} 
\usepackage{amssymb} 
\usepackage{stackengine} 
\usepackage{cite} 
\usepackage[numbers,sort&compress]{natbib}
\usepackage{tikz}
\usetikzlibrary{calc, positioning}
\usepackage{wrapfig}
\usepackage{subcaption}
\usetikzlibrary{decorations.pathreplacing}
\usepackage{svg}
\usepackage{balance}

\DeclarePairedDelimiter\norm{\lVert}{\rVert}%
\DeclarePairedDelimiter\abs{\lvert}{\rvert}

\def \towerWidth{1}
\def \towerHeigth{1.2}
\def \lowerPoint{0.2}
\def \higherPoint{0.7}
\def \antennaHeightScale{0.15}
\newcommand\codeTower{
	\coordinate (leftBottom) at (0, 0);
	\coordinate (rightBottom) at ($(leftBottom) + (\towerWidth, 0)$);
	\coordinate (peak) at ($(leftBottom) + (\towerWidth/2, \towerHeigth)$);
	\coordinate (peak2) at ($(peak) + (0, \antennaHeightScale*\towerHeigth)$);
	\coordinate (leftPoint1) at ($([scale=1-\lowerPoint]leftBottom)+([scale=\lowerPoint]peak)$);
	\coordinate (leftPoint2) at ($([scale=1-\higherPoint]leftBottom)+([scale=\higherPoint]peak)$);
	\coordinate (rightPoint1) at ($([scale=1-\lowerPoint]rightBottom)+([scale=\lowerPoint]peak)$);
	\coordinate (rightPoint2) at ($([scale=1-\higherPoint]rightBottom)+([scale=\higherPoint]peak)$);
	
	\draw[red] (leftBottom)  -- (peak) -- (rightBottom);
	\draw[red] (leftPoint1) -- (rightPoint2);
	\draw[red] (leftPoint2) -- (rightPoint1);
	\draw[red] [-stealth reversed] (peak) -- (peak2);
}
\def \userWidth{0.2}
\def \userHeigth{0.4}
\def \antennaWidth{0.05}
\def \antennaHeight{0.1}
\newcommand\codeUser{
	\coordinate (leftBottom) at (0, 0);
	\coordinate (rightTop) at ($(leftBottom) + (\userWidth, \userHeigth)$);
	\coordinate (antennaLeftBottom) at ($(rightTop) + (-\antennaWidth, 0)$);
	\coordinate (antennaRightTop) at ($(antennaLeftBottom) + (\antennaWidth, \antennaHeight)$);
	
	\draw [draw=green, fill=green] (leftBottom) rectangle (rightTop);
	\draw [draw=green] (antennaLeftBottom) rectangle (antennaRightTop);
}

\tikzset{
	zigzag/.style={to path={ -- ($(\tikztostart)!.55!-7:(\tikztotarget)$) -- ($(\tikztostart)!.45!7:(\tikztotarget)$) -- (\tikztotarget) \tikztonodes}}
}

\def\BibTeX{{\rm B\kern-.05em{\sc i\kern-.025em b}\kern-.08em
		T\kern-.1667em\lower.7ex\hbox{E}\kern-.125emX}}
\begin{document}

	\title{Pilot Contamination Aware Transformer for Downlink Power Control in Cell-Free Massive MIMO Networks}
	
	\author{Atchutaram~K.~Kocharlakota,~\IEEEmembership{Student Member,~IEEE,}
		Sergiy~A.~Vorobyov,~\IEEEmembership{Fellow,~IEEE,}
		and~ Robert~W.~Heath~Jr.,~\IEEEmembership{Fellow,~IEEE}
  \thanks{An initial part of this work was presented at 56th Asilomar Conference on Signals Systems, and Computers, Asilomar, CA, USA, Nov. 2022.}
\thanks{A. K. Kocharlakota and S. A. Vorobyov are with the Department of Information and Communications Engineering, Aalto University, PO Box 15400, 00076 Aalto, Finland. (e-mails: kameswara.kocharlakota@aalto.fi and sergiy.vorobyov@aalto.fi). A. K. Kocharlakota is also with Nokia, Espoo, Finland.}
\thanks{R. W. Heath Jr. is with the Department of Electrical and Computer Engineering, University of California San Diego, 9500 Gillman Dr, La Jolla, CA, US 92093. (e-mail: rwheathjr@ucsd.edu).}
\thanks{This material is based upon work supported in part by the National Science Foundation under Grant No. NSF-CCF-2435254.}
	}
	
	
	\maketitle
	\begin{abstract}
		Learning-based downlink power control in cell-free massive multiple-input multiple-output (CFmMIMO) systems offers a promising alternative to conventional iterative optimization algorithms, which are computationally intensive due to online iterative steps. Existing learning-based methods, however, often fail to exploit the intrinsic structure of channel data and neglect pilot allocation information, leading to suboptimal performance, especially in large-scale networks with many users. This paper introduces the pilot contamination-aware power control (PAPC) transformer neural network, a novel approach that integrates pilot allocation data into the network, effectively handling pilot contamination scenarios. PAPC employs the attention mechanism with a custom masking technique to utilize structural information and pilot data. The architecture includes tailored preprocessing and post-processing stages for efficient feature extraction and adherence to power constraints. Trained in an unsupervised learning framework, PAPC is evaluated against the accelerated proximal gradient (APG) algorithm, showing comparable spectral efficiency fairness performance while significantly improving computational efficiency. Simulations demonstrate PAPC’s superior performance over fully connected networks (FCNs) that lack pilot information, its scalability to large-scale CFmMIMO networks, and its computational efficiency improvement over APG. Additionally, by employing padding techniques, PAPC adapts to the dynamically varying number of users without retraining.
	\end{abstract}

	\IEEEpeerreviewmaketitle
	
	\section{Introduction}
	Base station (BS) coordination eliminates inter-cell interference and allows multi-user massive multiple-input multiple-output (MIMO) to serve users distributed over a large geographic area. Such coordination has been explored to increase per-user data rates and spectral efficiency (SE) of systems that are referred in the literature by different (not equivalent but having some incommon features) terms such as distributed multi-user MIMO antenna systems \cite{w2011multiuser,robert2013a}, cloud radio access networks (cloud RANs) \cite{y2010wireless,jun2015cloud}, and cell-free massive MIMO (CFmMIMO) systems \cite{ngo2017cellfree, manijeh2019maxmin, emil2020making, tugfe2021foundations} as used in this paper.
	
	To fully leverage the benefits of BS coordination, sophisticated pilot allocation and power control algorithms are essential. These algorithms face significant computational complexities due to the centralized signal processing tasks  \cite{g2019ubiquitous, alister2018cooperative, hamed2020performance}. Specifically, designing a downlink power control algorithm involves a large number of optimization parameters, posing a significant obstacle in the development of CFmMIMO infrastructure \cite{ngo2017cellfree, muhammad2021utility}. Addressing non-convex optimization problems is the core challenge, one that we confront in our paper.
	
	
	\subsection{Related literature}
	Power control algorithms in CFmMIMO typically employed various objective functions and second-order interior-point methods for optimization \cite{ngo2017cellfree, nguyen2017energy, ngo2018on, interdonato2019downlink}. For instance, the max-min fairness optimization was tackled using bisection search methods and second-order cone feasibility problems in \cite{ngo2017cellfree}. Similarly, energy efficiency maximization was addressed through successive convex approximation techniques in \cite{nguyen2017energy, ngo2018on}, which were also applied to the max-min fairness problem in \cite{interdonato2019downlink}. The scalability of second-order interior-point methods for power control in CFmMIMO systems remained a key challenge \cite{emil2020scalable, stefano2017cellfree, muhammad2021utility, c2022energy, interdonato2019downlink}. Two strategies were proposed to address this issue: user-centric approaches \cite{emil2020scalable, stefano2017cellfree, interdonato2019downlink} and first-order power control algorithms \cite{muhammad2021utility, c2022energy}, both offering reduced complexity. The user-centric approach enhanced efficiency by grouping BSs and setting large-scale fading coefficients between users and non-group BSs to zero, thereby reducing the computational burden. First-order methods reduced computational complexity by employing efficient solvers while operating on network-wide large-scale fading information without BS grouping, similar to second-order methods. The shift from second-order to first-order methods like accelerated projected gradient (APG) \cite{muhammad2021utility} aimed to improve scalability, but computational challenges persisted due to reliance on online iterative solvers, especially in large-scale networks. User-centric approaches also employed similar online iterative solvers.
	
	To overcome the computational challenges involved in the iterative solvers based power control, learning-based solutions were introduced. Trained offline to reduce complexity during inference, these methods map the input large-scale fading coefficients directly to output power control coefficients. This eliminates online iterative solvers, reducing computational complexity \cite{nuwanthika2021deep, nuwanthika2023unsupervised, rasoul2019unsupervisedlearning, akbar2022max, carmen2019uplink, yongshun2021deep, han2023user, mesquita2023decentralized, mostafa2022deep, xiaoqing2023deep, daesung2023learning, rasoul2020unsupervisedlearning, lirui2022downlink, rasoul2021unsupervised, lou2021deep, chongzheng2024joint, lou2022a, mahmoud2023learningbased, yongshun2023unsupervised, sucharita2023power}.
	
	Various learning-based solutions were proposed for uplink and downlink power control in CFmMIMO systems \cite{nuwanthika2021deep, nuwanthika2023unsupervised, rasoul2019unsupervisedlearning, akbar2022max, carmen2019uplink, yongshun2021deep, han2023user, mesquita2023decentralized, mostafa2022deep, xiaoqing2023deep, daesung2023learning, rasoul2020unsupervisedlearning, lirui2022downlink, rasoul2021unsupervised, lou2021deep, chongzheng2024joint, lou2022a}. A fully connected network (FCN) in an unsupervised learning setup was introduced in \cite{nuwanthika2021deep, nuwanthika2023unsupervised, rasoul2019unsupervisedlearning} for uplink power control with different objectives. A long short-term memory network was proposed in \cite{akbar2022max} under supervised learning, while FCNs were explored in \cite{carmen2019uplink, yongshun2021deep}. Reinforcement learning (RL) variants were proposed in \cite{han2023user, mesquita2023decentralized, mostafa2022deep, xiaoqing2023deep}. For downlink power control, unsupervised FCN solutions were studied in \cite{daesung2023learning, rasoul2020unsupervisedlearning}, and an RL solution was explored in \cite{lirui2022downlink}. Both uplink and downlink were addressed by unsupervised FCN in \cite{rasoul2021unsupervised}. Supervised convolutional neural network (CNN)-based solutions were proposed in \cite{lou2021deep, chongzheng2024joint}, and a graph neural network (GNN)-based solution for downlink appeared in \cite{lou2022a}. These algorithms commonly assumed orthogonal pilots and contamination-free environments.
	
	The studies in \cite{mahmoud2023learningbased, yongshun2023unsupervised, sucharita2023power} proposed deep learning solutions that considered pilot contamination scenarios during testing. A distributed unsupervised FCN for downlink power control was introduced in \cite{mahmoud2023learningbased}, while a similar solution for both uplink and downlink was presented in \cite{yongshun2023unsupervised}. In \cite{sucharita2023power}, a distributed RL solution for downlink power control was proposed. Although these models were tested with pilot contamination, they did not fully address pilot contamination during training or model design. Unlike traditional optimization-based power control algorithms, which utilize the pilot allocation information as input, they discard the pilot information.
	
	The effectiveness of neural network architectures for downlink power control depends on how they handle network-wide large-scale fading coefficients. FCN methods flatten the large-scale fading coefficient matrix (number of BSs $\times$ number of users), losing key associations between BSs and users. While CNN-based methods \cite{lou2021deep, chongzheng2024joint} preserve the matrix structure, their use is debatable since these matrices lack the localized clustering seen in natural images. In contrast, the GNN architecture in \cite{lou2022a} effectively leverages the structural information, providing a better solution for power control. None of the methods discussed, including the GNN, directly address pilot contamination.
	
	Pilot contamination remains a key challenge despite advancements in learning-based downlink power control methods. Ignoring pilot contamination limits the power control algorithms to CFmMIMO systems with a few number of users, restricting the number of users and impeding scalability due to the limited number of available orthogonal pilots. Unlike traditional optimization techniques, learning-based solutions lack explicit handling of pilot contamination, revealing a gap in addressing this issue while preserving the structure in the channel.
	
	\subsection{Contributions}
	Our paper introduces a novel Pilot contamination-Aware Power Control (PAPC) transformer neural network designed for downlink power control in a CFmMIMO network. PAPC leverages the large-scale fading coefficient matrix and a newly formulated matrix representation of pilot allocation information to map these inputs directly to power control coefficients. It is trained in an unsupervised fashion to maximize the empirical smoothed-minimum per-user spectral efficiency under power constraints, using modified transformer blocks with additional preprocessing and postprocessing modules. During inference, the model addresses the max-min fairness downlink power control problem. Training and testing are conducted in a time-division duplexed (TDD) CFmMIMO system with minimum mean square error channel estimation (MMSE) and matched filter downlink beamforming.\footnote{The model should be adaptable to other techniques and optimization objectives, though this is not directly tested in this work as it is outside of the scope.} The contributions of our paper are summarized as follows.
	\begin{itemize}
		\item \textbf{Attention mechanism for exploiting inter-user relationships:} The proposed PAPC transformer preserves the structural integrity of the network-wide two-dimensional large-scale fading coefficient matrix, avoiding the common pitfall of the flattening operation that can destroy crucial inter-user relationships. The attention mechanism in the transformer blocks of PAPC plays a vital role in extracting these relationships among the users. By doing so, it effectively learns to handle the dynamics of the propagation environment.
		
		\item \textbf{Modification of transformer blocks to incorporate pilot contamination matrix:} Without utilizing prior information from the pilot allocation algorithm, the power control algorithm's effectiveness is compromised. To address this, the standard transformer blocks are modified with a novel masking mechanism that incorporates the pilot contamination matrix into the neural network. This integration allows the PAPC to efficiently handle pilot contamination scenarios and provide enhanced inference accuracy in both contaminated and uncontaminated environments.
		
		\item \textbf{Enhancing accuracy with preprocessing and postprocessing stages:} The PAPC model's accuracy is enhanced by adding preprocessing and postprocessing stages. Preprocessing increases input dimensionality, allowing the model to learn richer features from each user's large-scale fading coefficients. Postprocessing adjusts the output power control coefficients back to the desired dimensionality while ensuring the compliance to power constraints.
		
		\item \textbf{Enabling adaptability:} The customized architecture of the postprocessing module, combined with padding techniques, allows the PAPC to handle varying numbers of users. This customization enables the model to adapt to different sizes of CFmMIMO without requiring a redesign or retraining.
		
		\item \textbf{Scalability through reduced hyperparameters:} The PAPC transformer's hyperparameters depend only on the number of BSs, unlike the FCN model, which also depends on the number of users. This reduction in hyperparameters enhances the scalability and efficiency in large CFmMIMO networks. The model's pilot contamination awareness also makes it suitable for handling a large number of users in these networks. The scalability of the PAPC model has been validated through extensive testing in large-scale CFmMIMO settings with up to $100$~APs and $80$~users.\footnote{We follow the characterization of large-scale CFmMIMO systems from \cite{yongshun2023unsupervised}, where a system is categorized as large-scale if $MK \geq 1000$.}
		
		\item \textbf{Benchmarking against advanced algorithms:} The performance of the PAPC model is validated against the network-wide first-order APG method from \cite{muhammad2021utility}. Known for its computational efficiency, the APG method converged about $100$ times faster in large-scale CFmMIMO systems, similar to those in this paper, compared to second-order methods. It serves as a strong benchmark for assessing the PAPC model. By targeting comparable performance to APG while significantly reducing computational complexity, the PAPC model demonstrates its potential for enabling large-scale CFmMIMO deployments.
	\end{itemize}
	Compared with prior work on centralized and distributed deep neural network (DNN)-based downlink power control algorithms \cite{daesung2023learning, rasoul2020unsupervisedlearning, rasoul2021unsupervised, mahmoud2023learningbased, yongshun2023unsupervised, lou2021deep, chongzheng2024joint, lou2022a}, PAPC not only leverages inter-user relationships within the network-wide large-scale fading coefficient matrix in a centralized manner but also accounts for pilot contamination scenarios by incorporating pilot allocation information. To the best of our knowledge, this is the first work to propose DNN-based pilot contamination-aware downlink power control algorithms in CFmMIMO.
	
	An earlier development in this direction was presented in a conference publication \cite{k2022attention}, but this paper significantly improves upon it by enhancing accuracy through improved feature extraction, introducing new preprocessing and postprocessing stages, and improved hyperparameter tuning.
 
	\subsection{Paper Organization}
	The paper is organized as follows. Section~\ref{System model} introduces the system model and provides the downlink SE expression for CFmMIMO networks. Section~\ref{DL PC} highlights challenges in designing a DNN for downlink power control, particularly addressing pilot contamination. Section~\ref{PAPC} describes the proposed PAPC transformer architecture. Section~\ref{sec:num_results} presents numerical evaluations and comparisons, and Section~\ref{conclusion} concludes the work with final observations.
	
	\subsection{Notation}
	The sets of real, positive real, and complex numbers are denoted by $\mathbb{R}$, $\mathbb{R}_{+}$, and $\mathbb{C}$, respectively. Matrices are represented using boldface capital letters, while vectors are indicated by boldface lowercase letters. For matrix and vector operations, the superscripts $(\cdot)^*$, $(\cdot)^{T}$, and $(\cdot)^H$ signify element-wise conjugate, transpose, and Hermitian transpose operations. Additionally, the symbol $\odot$ represents the Hadamard product, an element-wise product of two matrices or vectors. Element-specific operations are denoted as $a_i$ and $a_{i,j}$, representing the $i^{\text{th}}$ element of vector $\mathbf{a}$ and $(i,j)^{\text{th}}$ element of matrix $\mathbf{A}$, respectively, and  $\mathbf{1}_{K}$ represents a $K$ dimensional vector of ones. The function $\mathrm{ln}(\cdot)$ is used to denote the natural logarithm operation. For statistical notation, $\mathcal{N}_\mathcal{C}(m, \sigma^2)$ describes a circularly symmetric complex Gaussian random variable with mean vector $m$ and variance $\sigma^2$. The norm $\|\cdot\|$ indicates the $l_2$ norm of a vector, and $\mathbb{E}[\cdot]$ denotes the mathematical expectation of a random variable.
	
	\section{System Model} \label{System model}
	\def \scopeScale{0.7}
	\def \BW{8}
	\def \BH{8}
	\begin{figure}[!]
		\centering
		\begin{tikzpicture}[scale=0.9]
			\coordinate (B0) at (0, 0);
			\coordinate (B1) at ($(B0) + (1.15*\BW, 0)$);
			\coordinate (B2) at ($(B0) + (1.15*\BW, 1.15*\BH)$);
			\coordinate (B3) at ($(B0) + (0, 1.15*\BH)$);
			
			\coordinate (A0) at ($(B0) + (0.7*\BW, 0.9*\BH)$);
			\coordinate (A1) at ($(B0) + (0.4*\BW, 0.7*\BH)$);
			\coordinate (A2) at ($(B0) + (0.3*\BW, 0.2*\BH)$);
			\coordinate (A3) at ($(B0) + (1.0*\BW, 0.7*\BH)$);
			\coordinate (A4) at ($(B0) + (0.8*\BW, 0.2*\BH)$);
			\coordinate (A5) at ($(B0) + (0.1*\BW, 0.9*\BH)$);
			
			\coordinate (U0) at ($(B0) + (0.15*\BW, 0.75*\BH)$);
			\coordinate (U1) at ($(B0) + (0.65*\BW, 0.15*\BH)$);
			\coordinate (U2) at ($(B0) + (0.05*\BW, 0.45*\BH)$);
			\coordinate (U3) at ($(B0) + (0.75*\BW, 0.35*\BH)$);
			\coordinate (U4) at ($(B0) + (0.55*\BW, 0.45*\BH)$);
			\coordinate (U5) at ($(B0) + (0.35*\BW, 0.45*\BH)$);
			\coordinate (U6) at ($(B0) + (0.85*\BW, 0.55*\BH)$);
			\coordinate (U7) at ($(B0) + (1.00*\BW, 0.05*\BH)$);
			\coordinate (U8) at ($(B0) + (0.25*\BW, 0.15*\BH)$);
			\coordinate (U9) at ($(B0) + (0.40*\BW, 1.00*\BH)$);
			\coordinate (U10) at ($(B0) + (0.55*\BW, 0.85*\BH)$);
			\coordinate (U11) at ($(B0) + (0.90*\BW, 1.00*\BH)$);
			
			\coordinate (TAm) at ($(A0) + (0.4, -0.2)$);
			\coordinate (TA1) at ($(A5) + (0.4, -0.2)$);
			\coordinate (TAM) at ($(A4) + (0.4, -0.2)$);
			
			\coordinate (TUk) at ($(U10) + (0.1, -0.2)$);
			\coordinate (TU1) at ($(U8) + (0.1, -0.2)$);
			\coordinate (TUK) at ($(U4) + (0.1, -0.2)$);
			
			\coordinate (ARm) at ($(A0) + (\scopeScale*\towerWidth/2, \scopeScale*\towerHeigth+\scopeScale*\antennaHeightScale*\towerHeigth)$);
			
			\coordinate (UR1) at ($(U8) + (\userWidth, \userHeigth+\antennaHeight)$);
			\coordinate (URk) at ($(U10) + (\userWidth, \userHeigth+\antennaHeight)$);

			\coordinate (AR0) at ($(A0) + (\scopeScale*\towerWidth/2, \scopeScale*\towerHeigth+\scopeScale*\antennaHeightScale*\towerHeigth)$);
			\coordinate (AR1) at ($(A1) + (\scopeScale*\towerWidth/2, \scopeScale*\towerHeigth+\scopeScale*\antennaHeightScale*\towerHeigth)$);
			\coordinate (AR2) at ($(A2) + (\scopeScale*\towerWidth/2, \scopeScale*\towerHeigth+\scopeScale*\antennaHeightScale*\towerHeigth)$);
			\coordinate (AR3) at ($(A3) + (\scopeScale*\towerWidth/2, \scopeScale*\towerHeigth+\scopeScale*\antennaHeightScale*\towerHeigth)$);
			\coordinate (AR4) at ($(A4) + (\scopeScale*\towerWidth/2, \scopeScale*\towerHeigth+\scopeScale*\antennaHeightScale*\towerHeigth)$);
			\coordinate (AR5) at ($(A5) + (\scopeScale*\towerWidth/2, \scopeScale*\towerHeigth+\scopeScale*\antennaHeightScale*\towerHeigth)$);
			
			\coordinate (CS) at ($(B0) + (0.55*\BW, 0.6*\BH)$);
			\coordinate (CE) at ($(CS) + (0.8, 0.6)$);
			
			\coordinate (C1) at ($(CS) + (0.4, 0.6)$);
			\coordinate (C2) at ($(CS) + (0.8, 0.3)$);
			\coordinate (C3) at ($(CS) + (0.4, 0)$);
			\coordinate (C4) at ($(CS) + (0, 0.3)$);

			\begin{scope}[shift={(A0)}, scale=\scopeScale]
				\codeTower
			\end{scope}
			\begin{scope}[shift={(A1)}, scale=\scopeScale]
				\codeTower
			\end{scope}
			\begin{scope}[shift={(A2)}, scale=\scopeScale]
				\codeTower
			\end{scope}
			\begin{scope}[shift={(A3)}, scale=\scopeScale]
				\codeTower
			\end{scope}
			\begin{scope}[shift={(A4)}, scale=\scopeScale]
				\codeTower
			\end{scope}
			\begin{scope}[shift={(A5)}, scale=\scopeScale]
				\codeTower
			\end{scope}
			
			\node at (TAm) {\small BS $m$};
			\node at (TA1) {\small BS $1$};
			\node at (TAM) {\small BS $M$};

			\begin{scope}[shift={(U0)}]
				\codeUser
			\end{scope}
			\begin{scope}[shift={(U1)}]
				\codeUser
			\end{scope}
			\begin{scope}[shift={(U2)}]
				\codeUser
			\end{scope}
			\begin{scope}[shift={(U3)}]
				\codeUser
			\end{scope}
			\begin{scope}[shift={(U4)}]
				\codeUser
			\end{scope}
			\begin{scope}[shift={(U5)}]
				\codeUser
			\end{scope}
			\begin{scope}[shift={(U6)}]
				\codeUser
			\end{scope}
			\begin{scope}[shift={(U7)}]
				\codeUser
			\end{scope}
			\begin{scope}[shift={(U8)}]
				\codeUser
			\end{scope}
			\begin{scope}[shift={(U9)}]
				\codeUser
			\end{scope}
			\begin{scope}[shift={(U10)}]
				\codeUser
			\end{scope}
			\node at (TUk) {\small User $k$};
			\node at (TU1) {\small User $1$};
			\node at (TUK) {\small User $K$};
			\begin{scope}[shift={(U11)}]
				\codeUser
			\end{scope}
			
			\draw[-latex] (ARm) to[zigzag] node[anchor=south] {\small $\mathbf{g}_{mk}$ } (URk);
			\draw[-latex] (AR5) to[zigzag] node[anchor=north] {\small $\mathbf{g}_{1k}$ } (URk);
			\draw[-latex] (AR4) to[zigzag] node[anchor=west] {\small $\mathbf{g}_{Mk}$ } (URk);
			\draw[-latex] (AR4) to[zigzag] node[anchor=north] {\small $\mathbf{g}_{M1}$ } (UR1);
			\draw (B0) -- (B1) -- (B2) -- (B3) --cycle;
			\draw[color=blue] (CS) rectangle node {CP} (CE);
			
			\draw [dashed,color=blue] (C1) to [out=150,in=210] (A0);
			\draw [dashed,color=blue] (C2) to [out=20,in=210] (A3);
			\draw [dashed,color=blue] (C3) to [out=-160,in=0] (A2);
			\draw [dashed,color=blue] (C3) to [out=-80,in=180] (A4);
			\draw [dashed,color=blue] (C4) to [out=180,in=0] (A1);
			\draw [dashed,color=blue] (C4) to [out=200,in=-30] (A5);
			
		\end{tikzpicture}
		\caption{Illustration of the CFmMIMO system with $M$ distributed BSs, each equipped with $N$ antennas, serving $K$ single-antenna users under the coordination of a CP. The channel between BS $m$ and user $k$ is denoted by $\mathbf{g}_{mk}$.}
		\label{fig:CFmMIMO}
	\end{figure}
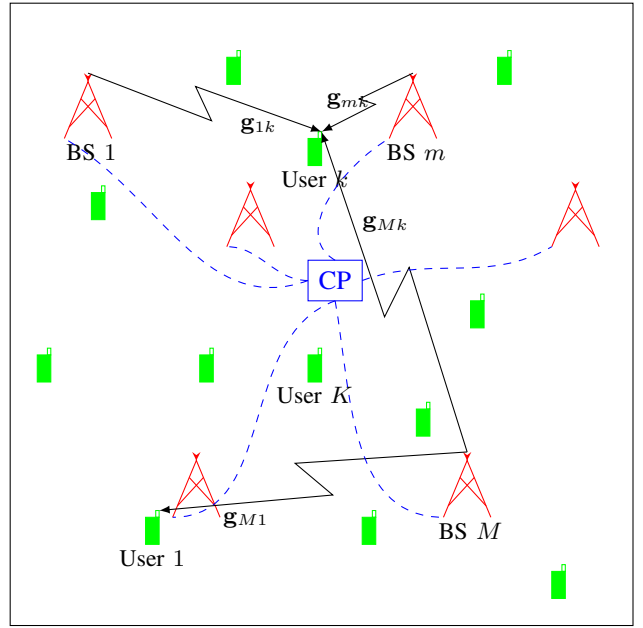
	Consider a fully synchronized time-division duplexed (TDD) CFmMIMO system where $M$ BSs, each with $N$ antennas, serve $K$ single-antenna users ($M > K$). All the BSs are connected to a central processor via a backhaul network and they simultaneously serve all users using common time-frequency resources, as depicted in Fig.~\ref{fig:CFmMIMO}.
	
	Assuming block-fading narrowband channels that remain constant over a coherence block of $\tau$ symbols, the uplink channel from the $k^{\text{th}}$ user to the $m^{\text{th}}$ BS is $\mathbf{g}_{mk} = \sqrt{\beta_{mk}} \mathbf{h}_{mk}$, where $\beta_{mk}$ is the large-scale fading coefficient and $\mathbf{h}_{mk} \in \mathbb{C}^{N}$ is the vector of small-scale fading coefficients with its elements independent and identically distributed (i.i.d.) as $\mathcal{N}_\mathcal{C}(0,1)$. The large-scale fading coefficients $\beta_{mk}, \; \forall m, k$ are constant over many coherence blocks, while the small-scale fading coefficients $\mathbf{h}_{mk}, \; \forall m, k$ change each block. The large-scale fading coefficients are assumed to be known at all the BS's and the central processor.
	
	During the uplink training phase, users transmit pilot sequences of length $\tau_p$ ($\tau_p \ll \tau$), where the $k^{\text{th}}$ user's pilot is $\sqrt{\tau_p} \boldsymbol{\psi}_k \in \mathbb{C}^{\tau_p}$ with $\norm{\boldsymbol{\psi}_k} = 1$. For $K>\tau_p$, due to limited number of orthogonal pilots, the sequences are reused, causing pilot contamination.
	
	The transmit signal-to-noise ratio (SNR) per pilot symbol is $\zeta_p$, and the additive white Gaussian noise (AWGN) at the $m^{\text{th}}$ BS is $\mathbf{Z}_{p,m} \in \mathbb{C}^{N \times \tau_p}$ with i.i.d. elements following the distribution $\mathcal{N}_\mathcal{C} (0,1)$. The received signal at the $m^{\text{th}}$ BS is
	\begin{align*}
		\mathbf{R}_{m} = \sqrt{\zeta_p \tau_p} \sum_{i=1}^{K} \mathbf{g}_{mi} \boldsymbol{\psi}_i^{T} + \mathbf{Z}_{p,m}.
	\end{align*}
	It is sytaightforward to find that the MMSE estimate of the channel between the $k^{\text{th}}$ user and the $m^{\text{th}}$ BS is
	\begin{align*}
		\hat{\mathbf{g}}_{mk} = \frac{\sqrt{\zeta_p \tau_p}\beta_{mk}}{1 + \zeta_p \tau_p \sum_{i=1}^{K} \beta_{mi} \abs{\boldsymbol{\psi}_{i}^{H} \boldsymbol{\psi}_{k}}^2} \mathbf{R}_{m} \boldsymbol{\psi}_{k}^{*},
	\end{align*}
	and the mean square value of each element in $\hat{\mathbf{g}}_{mk}$ is
	\begin{align*}
		\nu_{mk} = \mathbb{E}\left[\abs{\hat{\mathbf{g}}_{mk}[n]}^2\right] = \frac{\zeta_p \tau_p \beta_{mk}^2}{1 + \zeta_p \tau_p \sum_{i=1}^{K} \beta_{mi} \abs{\boldsymbol{\psi}_{i}^{H} \boldsymbol{\psi}_{k}}^2},
	\end{align*}
	which is uniform across all antenna elements $n$.
	
	In the downlink data transmission phase, let the $k^{\text{th}}$ user's data symbol be $c_k$ with $\mathbb{E}\left[\abs{c_k}^2\right] = 1$. Let $\zeta_d$ be the maximum transmit SNR per data symbol, and $\mu_{mi}$ be the power control coefficient for the signal to the $i^{\text{th}}$ user from the $m^{\text{th}}$ BS. The downlink transmit signal at the $m^{\text{th}}$ BS is
	\begin{align*}
		\mathbf{x}_{m} = \sqrt{\zeta_d} \sum_{i=1}^{K} \frac{\mu_{mi}}{\sqrt{\nu_{mi}}} \hat{\mathbf{g}}_{mi}^{*} c_i.
	\end{align*}
	Then, $\mathbb{E}\left[\norm{\mathbf{x}_{m}}^2\right] = \zeta_d N \sum_{i=1}^{K} \mu_{mi}^2$ represents the transmit power at the $m^{\text{th}}$ BS. The downlink received signal at the $k^{\text{th}}$ user is
	\begin{align*}
		r_k = \sum_{m=1}^{M} \mathbf{g}_{mk}^{T} \mathbf{x}_{m} + z_{d,k}.
	\end{align*}
	where $z_{d,k} \sim \mathcal{N}_\mathcal{C}(0, 1)$ is the AWGN noise.
	
	Define a vector of power control coefficients for the $k^{\text{th}}$ user as $\bar{\boldsymbol{\mu}}_{k}~=~[\mu_{1k}, \ldots, \mu_{Mk}]^{T}$, a diagonal matrix $\mathbf{D}_{k}$ as $\mathbf{D}_{k} = \operatorname{diag}(\sqrt{\beta_{1k}}, \ldots, \sqrt{\beta_{Mk}})$, and a vector $\boldsymbol{\nu}_{ik} \in \mathbb{R}^{M}$ such that
	\begin{align*}
		\boldsymbol{\nu}_{ik}[m] = \abs{\boldsymbol{\psi}_{k}^{T} \boldsymbol{\psi}_{i}^{*}} \sqrt{\nu_{mi}} \frac{\beta_{mk}}{\beta_{mi}}.
	\end{align*}
	Using the use-and-then-forget bounding technique, the signal-to-interference-plus-noise ratio (SINR) for the $k^{\text{th}}$ user is \cite{muhammad2021utility, ngo2018on}
	\begin{align}
		\gamma_k = \frac{\zeta_d \left( \bar{\boldsymbol{\mu}}_{k}^{T} \boldsymbol{\nu}_{kk} \right)^2}{\sum\limits_{\substack{i=1 \\ i \neq k}}^{K}\zeta_d \left( \bar{\boldsymbol{\mu}}_{i}^{T} \boldsymbol{\nu}_{ik} \right)^2+ \frac{\zeta_d}{N} \sum\limits_{i=1}^{K}\norm{\mathbf{D}_{k} \bar{\boldsymbol{\mu}}_{i}}^{2} + \frac{1}{N^2}}, \label{sinr}
	\end{align}
	and a lower bound on the downlink SE (bits/s/Hz) is
	\begin{align}
		\text{SE}_k = \left(1- \frac{\tau_p}{\tau}\right) \log_2(1 + \gamma_k) . 
		\label{SE}
	\end{align}
	
	Define the large-scale fading coefficients matrix as $\mathbf{B} \in \mathbb{R}_{+}^{M \times K}$. The element at position $(m, k)$ of $\mathbf{B}$ is $\beta_{mk}$. Let $\boldsymbol{\mu}_{m} = [\mu_{m1}, \ldots, \mu_{mK}]^{T}$ denote the power control coefficients at the $m^{\text{th}}$ BS. Define the matrix of all power control coefficients as
	\begin{align}
		\mathbf{M} = 
		\begin{bmatrix}
			\mu_{11} & \cdots & \mu_{1K}\\
			\vdots & \ddots & \vdots \\
			\mu_{M1} & \cdots & \mu_{MK}
		\end{bmatrix}
		=
		\begin{bmatrix}
			\boldsymbol{\mu}_{1}^{\intercal}\\
			\vdots\\
			\boldsymbol{\mu}_{M}^{\intercal}
		\end{bmatrix}
		= 
		\begin{bmatrix}
			\bar{\boldsymbol{\mu}}_{1} & \cdots & \bar{\boldsymbol{\mu}}_{K}
		\end{bmatrix}. \label{matrix_M}
	\end{align}
	Then the SE is a function of power control coefficients and pilots, while the large-scale fading coefficients are considered fixed parameters that characterize the function. Given the large-scale fading coefficients (matrix $\mathbf{B}$), the central processor is responsible to perform optimal pilot allocation (compute $K$ pilots, $\boldsymbol{\psi}$'s) and power control (compute matrix $\mathbf{M}$).
	
	\section{Challenges in DNN-Based Power Control} \label{DL PC}
	\subsection{Need for Learning-based Approaches}
	Max-min fairness optimization approaches in downlink power control aim to maximize the minimum SE across users.  These methods target maximizing the SE of the worst-performing user to ensure fairness, while managing power distribution and pilot allocation across the network. The APG algorithm \cite{ngo2017cellfree} is the state-of-the-art solution for power control in large-scale CFmMIMO systems, efficiently handling power control through a first-order accelerated method. Power control is subject to constraints that limit the total transmitted power per BS, expressed using the set:
	\begin{equation}
		\mathcal{S} = \left\{ \mathbf{M} \;\middle|\;
		\begin{aligned}
			& \mathbf{M} \ge \mathbf{0}, \\ 
			& \norm{\boldsymbol{\mu}_{m}}^{2} \le \frac{1}{N}, \\
			& \forall m \in \{1, \dots, M\}
		\end{aligned}
		\right\}.
	\end{equation}
	
	While traditional methods rely on iterative solvers, learning-based approaches offer an alternative to mitigate the computational costs \cite{nuwanthika2021deep, nuwanthika2023unsupervised, rasoul2019unsupervisedlearning, akbar2022max, carmen2019uplink, yongshun2021deep, han2023user, mesquita2023decentralized, mostafa2022deep, xiaoqing2023deep, daesung2023learning, rasoul2020unsupervisedlearning, lirui2022downlink, rasoul2021unsupervised, lou2021deep, chongzheng2024joint, lou2022a, mahmoud2023learningbased, yongshun2023unsupervised, sucharita2023power}. In supervised learning, large datasets are used to train DNN models that approximate the unknown function mapping inputs of the iterative algorithms to the output power control coefficients. Despite lacking a closed-form expression, learning-based methods can efficiently learn this mapping through extensive training.
	
	Unsupervised learning, unlike supervised methods, does not rely on reference solutions generated by iterative algorithms. Both approaches perform function approximation, but in unsupervised learning, the DNN  model approximates a different unknown function: the one that models the relationship between the input of the model and power control coefficients to directly maximize the minimum SE. The unsupervised learning method, which is used in this paper, also reduces the need for large amounts of labeled data.
	
	\subsection{Essential Inputs for Power Control Optimization} \label{DNN_inputs}
	Efficient function approximation using DNN necessitates a thorough understanding of the structural complexities in $\mathbf{B}$ and the interdependencies between each user's SE and key variables, including pilot contamination dynamics and the matrices $\mathbf{B}$ and $\mathbf{M}$.
	
	Consider the following two terms contributing to $2K^2$ variables, with indices $i$ and $k$ ranging from $1$ to $K$:
	\begin{align}
		\bar{\boldsymbol{\mu}}_{i}^{T}\boldsymbol{\nu}_{ik} &= \left|\boldsymbol{\psi}_{k}^{T} \boldsymbol{\psi}_{i}^{*}\right| \sum_{m=1}^{M}\sqrt{\nu_{mi}} \mu_{mi}\frac{\beta_{mk}}{\beta_{mi}} \label{key_term1}
	\end{align}
	and
	\begin{align}
		\norm{\mathbf{D}_{k} \bar{\boldsymbol{\mu}}_{i}}^{2} &= \sum_{m=1}^{M}\mu^{2}_{mi}\beta_{mk} \label{key_term2}.
	\end{align}
	From \eqref{sinr} and \eqref{SE}, the SE expressions for all users are composed of these $2K^2$ variables. Thus, their interplay determines $\underset{1 \leq k \leq K}{\min} \, \text{SE}_k$. To maximize $\underset{1 \leq k \leq K}{\min} \, \text{SE}_k$, the DNN should learn to generate the power control coefficient matrix $\mathbf{M}$, which determines the $2K^2$ control variables directly influencing the minimum SE.
	
	Let $\boldsymbol{\Phi} \in \mathbb{R}^{K \times K}$ represent the pilot allocation information, where $\abs{\boldsymbol{\psi}_{i}^{T} \boldsymbol{\psi}_{j}^{*}}^2$ is the element at $(i, j)$ of $\boldsymbol{\Phi}$, i.e., $\phi_{i,j}$. From \eqref{key_term1}, incorporating $\boldsymbol{\Phi}$ into the DNN is essential. Without it, the DNN cannot generate $\mathbf{M}$ that can effectively control the $K^2$ variables associated with $\bar{\boldsymbol{\mu}}_{i}^{T}\boldsymbol{\nu}_{ik}$. Similarly, equations \eqref{key_term1} and \eqref{key_term2} underscore the importance of including $\mathbf{B}$ as input, while equation \eqref{key_term1} emphasizes the need to preserve inter-user relationships between its columns, especially the element-wise ratios. Therefore, the DNN should incorporate both $\boldsymbol{\Phi}$ and $\mathbf{B}$ as inputs while preserving the structure of $\mathbf{B}$.
	
	\subsection{Optimization Problem for DNN Training}
	Assuming pilot allocation is performed prior to power control, and $\boldsymbol{\Phi}$ is available as input to the power control algorithm, define an arbitrary DNN, $\mathbf{F}_{\text{D}}(\cdot, \cdot; \mathbf{W})$, that takes $\boldsymbol{\Phi}$ and $\mathbf{B}$ as inputs, $\mathbf{M} = \mathbf{F}_{\text{D}}(\boldsymbol{\Phi}, \mathbf{B}; \mathbf{W})$, where $\mathbf{W}$ represents the trainable parameters.
	
	Since $\boldsymbol{\Phi}$ is known, unlike in Sec.~\ref{System model}, the SE is now a function solely of power control coefficients, while the large-scale fading coefficients  and pilots are considered fixed parameters that characterize the function.
	\def \heightOffset{0}
	\def \lineWidth{1.25}
	\def \lineHeight{2.5}
	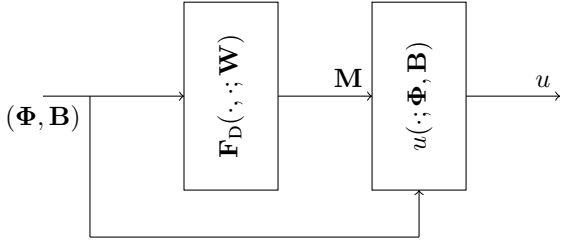
\begin{figure}
		\centering
		\begin{tikzpicture}[scale=1]
			
			\coordinate (A) at (0, 0);
			\coordinate (B) at ($(A) + (\lineWidth/2, 0)$);
			\coordinate (C) at ($(B) + (0, -0.75*\lineHeight)$);
			\coordinate (D) at ($(B) + (\lineWidth, 0)$);
			\coordinate (E) at ($(D) + (0, -\lineHeight/2)$);
			\coordinate (F) at ($(E) + (\lineWidth, \lineHeight)$);
			\coordinate (G) at ($(F) + (0, -\lineHeight/2)$);
			\coordinate (H) at ($(G) + (\lineWidth, 0)$);
			\coordinate (I) at ($(C) + (3.5*\lineWidth, 0)$);
			\coordinate (J) at ($(I) + (0, 0.25*\lineHeight)$);
			
			\coordinate (Q) at ($(H) + (0, -0.5*\lineHeight)$);
			\coordinate (R) at ($(Q) + (\lineWidth, \lineHeight)$);
			
			\coordinate (S) at ($(R) + (0, -\lineHeight/2)$);
			\coordinate (T) at ($(S) + (\lineWidth, 0)$);
			
			
			\draw (A) node[below] {$\left(\boldsymbol{\Phi}, \mathbf{B}\right)$}-- (B);
			\draw (B) -- (C);
			
			\draw[->] (B) -- (D);
			\draw (E) rectangle (F) node[pos=.5,align=center,rotate=90] {$\mathbf{F}_{\text{D}}(\cdot, \cdot; \mathbf{W})$};
			\draw[->] (G) -- (H)  node[anchor=south east] {$\mathbf{M}$};
			
			\draw (C) -- (I);
			\draw[->] (I) -- (J);
			
			\draw (Q) rectangle (R) node[pos=.5,rotate=90] {$u(\cdot; \boldsymbol{\Phi}, \mathbf{B})$};
			\draw[->] (S) -- (T)  node[anchor=south east] {$u$};

		\end{tikzpicture}
		\caption{Diagram of the unsupervised learning framework for downlink power control in CFmMIMO. The DNN, $F_D(\cdot, \cdot; W)$, takes the large-scale fading coefficient matrix $\mathbf{B}$ and the pilot allocation information $\boldsymbol{\Phi}$ as inputs to generate the power control coefficients. The DNN is trained to generate these coefficients to minimize the average of the utility function, $u(\cdot; \boldsymbol{\Phi}, \mathbf{B})$.}
		\label{fig:unsup_setup}
	\end{figure}
	
	Assuming fixed BS placement and a path loss model, generate a set of large-scale fading coefficient matrices $\{\mathbf{B}^{[p]} \in \mathbb{R}_{+}^{M \times K} \mid p = 1, \dots, P\}$ for $P$ random user placements, which are then used by the pilot allocation algorithm to produce $\{\boldsymbol{\Phi}^{[p]} \in \mathbb{R}_{+}^{M \times K} \mid p = 1, \dots, P\}$.
	
	Similar to the APG method in \cite{ngo2017cellfree}, consider a soft-minimum utility function with smoothening parameter $\lambda$ as:
	\begin{equation}
		u(\mathbf{M}; \boldsymbol{\Phi}, \mathbf{B}) = - \frac{1}{\lambda} \ln\left(\frac{1}{K} \sum_{k=1}^{K} e^{-\lambda \text{se}_k(\mathbf{M}; \boldsymbol{\Phi}, \mathbf{B})}\right), \label{utility_fn}
	\end{equation}
	and the utility function for each sample as
	\begin{align}
		u^{[p]}(\mathbf{W}) = u\left(\mathbf{F}_{\text{D}}(\boldsymbol{\Phi}^{[p]}, \mathbf{B}^{[p]}; \mathbf{W}); \boldsymbol{\Phi}^{[p]}, \mathbf{B}^{[p]}\right).
	\end{align}
	The DNN training finds the optimal weights $\mathbf{W}_{\text{opt}}$:
	\begin{equation}
		\mathbf{W}_{\text{opt}} = \begin{aligned}
			& \underset{\mathbf{W}}{\text{argmax}} & \frac{1}{P}\sum_{p=1}^{P} u^{[p]}(\mathbf{W}).
		\end{aligned} \label{opt_problem_empirical}
	\end{equation}
	During inference, this aims to replace expensive solvers for the max-min fairness problem, seeking the mapping $\mathbf{F}_{\text{D}}(\boldsymbol{\Phi}, \mathbf{B}; \mathbf{W}_{\text{opt}})$ that achieves similar performance. The unsupervised learning setup is depicted by Fig.~\ref{fig:unsup_setup}.
	
	\subsection{Towards an Efficient DNN Design}
	The structure of the DNN, $\mathbf{F}_{\text{D}}(\cdot, \cdot; \mathbf{W})$, plays a crucial role in achieving comparable performance to iterative solvers. Its design is essential for effectively handling the function approximation task. While FCN models are favored for their universal approximation abilities, they require flattened $\mathbf{B}$, leading to convergence issues in large-scale systems. They also fail to incorporate $\boldsymbol{\Phi}$, limiting their use to smaller systems. These limitations underscore the need for alternative architectures that preserve structural integrity of $\mathbf{B}$ and effectively address pilot contamination, enabling scalability.
	
	\section{The PAPC Transformer} \label{PAPC}
	\subsection{Overview of GPT}\label{sec:GPT-overview}
	Introduced in \cite{ashish2017attention}, the transformer architecture has significantly impacted the field of natural language processing (NLP), enabling advancements in tasks such as machine translation, text summarization, and sentiment analysis. Among the architectures inspired by this model, generative pre-transformers (GPT) \cite{alec2018improving} stand out for their impact on language generation. While the proposed neural network differs in both purpose and training methodology from GPT, it shares significant structural similarities. This subsection provides a broad overview of GPT to lay the foundation for the proposed network's design.
	
	GPT’s preprocessing involves tokenization, embedding, and positional encoding. The input text is broken into tokens (words or subwords), transformed into numerical vectors through embedding to capture semantic features, and then augmented with positional encoding to preserve the sequence order. These encoded vectors are fed into GPT, which uses its attention mechanisms to extract context-aware features and perform tasks like next-token prediction.
	
	At the core of GPT is the masked multi-head attention (MMHA) mechanism, which captures dynamic \textit{interrelationships} among tokens by focusing on relevant parts of the input sequence. Masking enforces causality, restricting the model's attention to the current and preceding tokens, supporting GPT’s autoregressive nature. The multi-head structure allows simultaneous exploration of token dependencies from various perspectives, enhancing the model’s ability to generate coherent text.
	
	The GPT architecture consists of multiple blocks, each integrating MMHA with normalization layers and feed-forward networks. These blocks sequentially process the input, refining token representations to predict the next token accurately.
	
	Although GPT was originally developed for language tasks, its key features—attention and masking—can be adapted to non-NLP applications such as power control. In this context, the attention mechanism can model inter-user relationships from input data like $\mathbf{B}$, and the masking mechanism can be adjusted to incorporate $\boldsymbol{\Phi}$ as input.
	
	\subsection{PAPC for Downlink Power Control}
	This paper proposes the PAPC transformer model, designed to address the challenges of DNN-based downlink power control in CFmMIMO. Building upon the foundational principles of transformer architectures like GPT, PAPC effectively models the interactions between the columns of the large-scale fading matrix $\mathbf{B}$, representing the dynamic inter-user channel dependencies. By learning these interdependencies, the model computes power control coefficients. It incorporates pilot allocation matrix $\boldsymbol{\Phi}$ using customized masking functionality. The PAPC model and its core components are outlined here.
	
	\subsubsection{Layer Normalization}
	A customized layer normalization technique is introduced as a fundamental unit of the PAPC transformer, applied at several stages throughout the architecture.
	
	Layer normalization consists of two key steps: normalization and feature-specific scaling and shifting. Let $\mathbf{C}$ represent the input matrix, where each row is a feature vector of length $\ddot{M}$ corresponding to a user.
	
	First, the normalization process computes the mean and standard deviation across all elements of $\mathbf{C}$. Each element is then adjusted by subtracting the mean and dividing by the standard deviation, resulting in $\bar{\mathbf{C}}$.
	
	\def \heightOffset{0}
	\def \lineWidth{1}
	\def \lineHeight{2.2}
	\def \circleOffsetX{5}
	\def \circleOffsetY{4.5}
	\def \NumRectangles{3}
	\def \DeltaX{0.1}
	\def \DeltaY{0.1}
	\def\TextOffset{0.5}
	
	\begin{figure}
		\centering
		\begin{tikzpicture}[scale=1]
			
			\coordinate (A) at (0, 0);
			\coordinate (B) at ($(A) + (\lineWidth/2, 0)$);
			\coordinate (C) at ($(B) + (0, -\lineHeight/2)$);
			\coordinate (D) at ($(C) + (\lineWidth, \lineHeight)$);
			\coordinate (E) at ($(D) + (0, -\lineHeight/2)$);
			\coordinate (F) at ($(E) + (\lineWidth/2, 0)$);
			\coordinate (G) at ($(F) + (0, -\lineHeight/2)$);
			\coordinate (H) at ($(G) + (\lineWidth, \lineHeight)$);
			\coordinate (I) at ($(H) + (0, -\lineHeight/2)$);
			\coordinate (J) at ($(I) + (\lineWidth/2, 0)$);
			\coordinate (K) at ($(J) + (0, -\lineHeight/2)$);
			\coordinate (L) at ($(K) + (\lineWidth, \lineHeight)$);
			\coordinate (M) at ($(L) + (0, -\lineHeight/2)$);
			\coordinate (N) at ($(M) + (\lineWidth/2, 0)$);
			
			\coordinate (X) at ($(A) + (\lineWidth/4, -\lineHeight*0.75)$);
			\coordinate (Y) at ($(X) + (\lineWidth*5.3, \lineHeight*1.5)$);

			\draw[->, line width=0.3mm] (A) node[left] {$\mathrm{ln}{\mathbf{B}^{T}}$}-- (B);
			\draw[fill=white] (C) rectangle (D) node[pos=.5,align=center,rotate=90] {LayerNorm};
			\draw[->, line width=0.3mm] (E) -- (F);
			
			\draw[fill=white] (G) rectangle (H) node[pos=.5,align=center,rotate=90] {Linear};

			\draw[->, line width=0.3mm] (I) -- (J);
			\draw[fill=white] (K) rectangle (L) node[pos=.5,align=center,rotate=90] {LayerNorm};
			\draw[->, line width=0.3mm] (M) -- (N) node[right] {$\mathbf{Z}^{(0)}$};

		\end{tikzpicture}
		\caption{Preprocessing stage of the CFmMIMO power control model. The input matrix $\mathbf{B}$ is transposed, log-transformed, and mapped to a higher-dimensional representation $\mathbf{Z}^{(0)}$ for use in subsequent transformer blocks.}
		\label{fig:preprocessing}
	\end{figure}
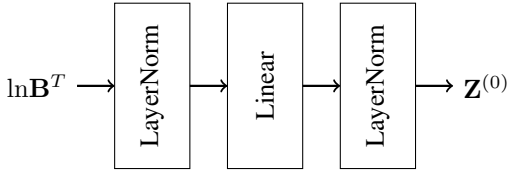
	
	\begin{figure*}
		\centering
		\begin{subfigure}[b]{0.35\textwidth}
			\centering
			\begin{tikzpicture}[scale=1]
				\def \heightOffset{0}
				\def \lineWidth{2}
				\def \lineHeight{1}
				\def \nnHeight{5}
				\def \nnWidth{4}
				\def \circleOffsetX{5.5}
				\def \circleOffsetY{5}
				\def \NumRectangles{3}
				\def \DeltaX{0.3}
				\def \DeltaY{0.1}
				\def\TextOffset{0.6}
				
				\centering
				
				\coordinate (A) at (0, 0);
				\coordinate (AShift) at ($(A) - (2*\DeltaX,2*\DeltaY)$);
				\coordinate (MidShift) at ($(AShift) + (0, \lineHeight/2)$);
				\coordinate (BShift) at ($(MidShift) + (0, \lineHeight/2)$);
				
				\coordinate (Start) at (2.5, 3.8);
				\coordinate (End) at (2.9, 4);
				
				\coordinate (MidArc) at ($(Start)!0.5!(End)$);
				
				\draw[line width=0.3mm] (AShift) node[below] {$\mathbf{X}$} -- (MidShift);
				\draw[line width=0.3mm] (MidShift) -- (BShift);
				
				\foreach \i in {1,...,\NumRectangles}{
					\pgfmathsetmacro{\opacityValue}{min(1, 0.1+0.2*\i)} 
					\pgfmathsetmacro{\fillIntensity}{min(100, 30*\i)} 
					\ifnum\i=3
					\pgfmathsetmacro{\opacityValue}{1} 
					\pgfmathsetmacro{\fillIntensity}{100} 
					\fi
					\coordinate (AShift) at ($(A) - (\i*\DeltaX,\i*\DeltaY) + (\DeltaX,\DeltaY)$);
					\coordinate (Mid) at ($(AShift) + (0, \lineHeight/2)$);
					\coordinate (B) at ($(Mid) + (0, \lineHeight/2)$);
					\coordinate (C) at ($(B) + (-\lineWidth, 0)$);
					\coordinate (D) at ($(B) + (\lineWidth, 0)$);
					\coordinate (E) at ($(C) + (0, \lineHeight)$);
					\coordinate (F) at ($(B) + (0, \lineHeight)$);
					\coordinate (G) at ($(D) + (0, \lineHeight)$);
					
					\coordinate (H) at ($(E) + (-1.3*\lineWidth/4, 0)$);
					\coordinate (I) at ($(H) + (1.3*\lineWidth/2, \lineHeight*0.6)$);
					\coordinate (J) at ($(F) + (-1.3*\lineWidth/4, 0)$);
					\coordinate (K) at ($(J) + (1.3*\lineWidth/2, \lineHeight*0.6)$);
					\coordinate (L) at ($(G) + (-1.3*\lineWidth/4, 0)$);
					\coordinate (M) at ($(L) + (1.3*\lineWidth/2, \lineHeight*0.6)$);
					
					\coordinate (N) at ($(I) + (-1.3*\lineWidth/4, 0)$);
					\coordinate (O) at ($(K) + (-1.3*\lineWidth/4, 0)$);
					\coordinate (P) at ($(M) + (-1.3*\lineWidth/4, 0)$);
					
					\coordinate (Q) at ($(N) + (0, \lineHeight)$);
					\coordinate (R) at ($(O) + (0, \lineHeight)$);
					\coordinate (S) at ($(P) + (0, \lineHeight)$);
					
					\coordinate (T) at ($(Q) + (-\lineWidth/2, 0)$);
					\coordinate (U) at ($(T) + (3*\lineWidth, \lineHeight*3/4)$);
					\coordinate (TT) at ($(T) + (0, \lineHeight*3/8)$);
					\coordinate (UU) at ($(TT) + (-\lineWidth/4, 0)$);
					
					\coordinate (V) at ($(U) + (-\lineWidth*1.5, 0)$);
					\coordinate (W) at ($(V) + (0, \lineHeight)$);
					\coordinate (X) at ($(W) + (-\lineWidth/2, 0)$);
					\coordinate (Y) at ($(X) + (\lineWidth, \lineHeight/2)$);
					
					\coordinate (Z) at ($(Y) + (-\lineWidth/2, 0)$);
					\coordinate (AA) at ($(Z) + (0, \lineHeight)$);
					\coordinate (BB) at ($(AA) + (-1.3*\lineWidth/4, 0)$);
					\coordinate (CC) at ($(BB) + (1.3*\lineWidth/2, \lineHeight*0.6)$);
					
					\coordinate (DD) at ($(CC) + (-1.3*\lineWidth/4, 0)$);
					\coordinate (EE) at ($(DD) + (0, \lineHeight)$);
					
					\coordinate (XX) at ($(B) + (-\lineWidth*1.75, -\lineHeight*1.2)$);
					\coordinate (YY) at ($(XX) + (3.5*\lineWidth, 9.5*\lineHeight)$);
					
					\draw[draw opacity=\opacityValue, line width=0.3mm]  (MidShift) -- (Mid);
					\draw[draw opacity=\opacityValue, line width=0.3mm]  (Mid) -- (B);
					\draw[draw opacity=\opacityValue, line width=0.3mm]  (B) -- (C);
					\draw[draw opacity=\opacityValue, line width=0.3mm]  (B) -- (D);
					\ifnum\i=3
					\draw[draw opacity=\opacityValue, ->, line width=0.3mm]  (C) -- (E);
					\draw[draw opacity=\opacityValue, ->, line width=0.3mm]  (B) -- (F);
					\draw[draw opacity=\opacityValue, ->, line width=0.3mm]  (D) -- (G);
					\draw[rounded corners=2mm, fill=cyan!\fillIntensity!white, draw=black, fill opacity=\opacityValue, draw opacity=\opacityValue]  (H) rectangle (I) node[midway] {Linear};
					\draw[rounded corners=2mm, fill=cyan!\fillIntensity!white, draw=black, fill opacity=\opacityValue, draw opacity=\opacityValue]  (J) rectangle (K) node[midway] {Linear};
					\draw[rounded corners=2mm, fill=cyan!\fillIntensity!white, draw=black, fill opacity=\opacityValue, draw opacity=\opacityValue]  (L) rectangle (M) node[midway] {Linear};
					
					\draw[draw opacity=\opacityValue, ->, line width=0.3mm] (N) node[left, xshift=-0.1cm, yshift=0.5cm] {Q}-- (Q);
					\draw[draw opacity=\opacityValue, ->, line width=0.3mm] (O) node[left, xshift=-0.1cm, yshift=0.5cm] {K}-- (R);
					\draw[draw opacity=\opacityValue, ->, line width=0.3mm] (P) node[left, xshift=-0.1cm, yshift=0.5cm] {V}-- (S);
					\draw[rounded corners=2mm, fill=lightgray!\fillIntensity!white, draw=black, fill opacity=\opacityValue, draw opacity=\opacityValue]  (T) rectangle (U) node[midway, align=center, text width=6cm] {Scaled dot-product self-attention};
					\draw[->, line width=0.3mm] (UU) node[left] {$\boldsymbol{\Phi}$}-- (TT);
					\else
					\draw[draw opacity=\opacityValue, line width=0.3mm]  (C)  -- (E);
					\draw[draw opacity=\opacityValue, line width=0.3mm]  (B)  -- (F);
					\draw[draw opacity=\opacityValue, line width=0.3mm]  (D)  -- (G);
					\draw[rounded corners=2mm, fill=cyan!\fillIntensity!white, draw=black, fill opacity=\opacityValue, draw opacity=\opacityValue]  (H) rectangle (I);
					\draw[rounded corners=2mm, fill=cyan!\fillIntensity!white, draw=black, fill opacity=\opacityValue, draw opacity=\opacityValue]  (J) rectangle (K);
					\draw[rounded corners=2mm, fill=cyan!\fillIntensity!white, draw=black, fill opacity=\opacityValue, draw opacity=\opacityValue]  (L) rectangle (M);
					
					\draw[draw opacity=\opacityValue, line width=0.3mm] (N)-- (Q);
					\draw[draw opacity=\opacityValue, line width=0.3mm] (O)-- (R);
					\draw[draw opacity=\opacityValue, line width=0.3mm] (P)-- (S);
					\draw[rounded corners=2mm, fill=lightgray!\fillIntensity!white, draw=black, fill opacity=\opacityValue, draw opacity=\opacityValue]  (T) rectangle (U);
					\fi
					
					\draw[draw opacity=\opacityValue, ->, line width=0.3mm]  (V) -- (W);
				}
				
				\draw [decorate,decoration={brace,amplitude=10pt,mirror,raise=4pt},yshift=0pt, line width=0.3mm]	(2.3,3.5) -- (3.0,3.7);
				\node at (3.3,2.9) {$H$ heads}; 
				
				\draw[rounded corners=2mm, fill=yellow!30!white, draw=black] (X) rectangle (Y) node[midway] {Concat};
				
				\draw[->, line width=0.3mm] (Z) -- (AA);
				\draw[rounded corners=2mm, fill=cyan, draw=black] (BB) rectangle (CC) node[midway] {Linear};
				\draw[->, line width=0.3mm] (DD) -- (EE) node[above] {$\mathbf{Y}$};
			\end{tikzpicture}
			\caption{Overview of MMHA with multi-head attention and output combination.}
			\label{fig:MMHA}
		\end{subfigure}
		\hfill 
		\begin{subfigure}[b]{0.55\textwidth}
			\centering
			\begin{tikzpicture}
				\def \heightOffset{0}
				\def \lineWidth{3}
				\def \lineHeight{1}
				\def \nnHeight{5}
				\def \nnWidth{4}
				\def \circleOffsetX{5.5}
				\def \circleOffsetY{5}
				\def \NumRectangles{3}
				\def \DeltaX{0.3}
				\def \DeltaY{0.1}
				\def\TextOffset{0.6}
				
				\coordinate (A) at (0, 0);
				\coordinate (B) at ($(A) + (\lineWidth/3, 0)$);
				\coordinate (C) at ($(B) + (\lineWidth/2, 0)$);
				\coordinate (D) at ($(A) + (0, \lineHeight)$);
				\coordinate (E) at ($(B) + (0, \lineHeight)$);
				\coordinate (F) at ($(D) + (-\lineWidth/3, 0)$);
				\coordinate (G) at ($(F) + (\lineWidth, \lineHeight/2)$);
				\coordinate (H) at ($(G) + (-\lineWidth/2, 0)$);
				\coordinate (I) at ($(H) + (0, \lineHeight)$);
				\coordinate (J) at ($(I) + (-\lineWidth/2, 0)$);
				\coordinate (K) at ($(J) + (\lineWidth, \lineHeight/2)$);
				\coordinate (L) at ($(K) + (-\lineWidth/2, 0)$);
				\coordinate (M) at ($(L) + (0, \lineHeight)$);
				\coordinate (N) at ($(M) + (-\lineWidth/2, 0)$);
				\coordinate (Q) at ($(N) + (0, \lineHeight/4)$);
				\coordinate (P) at ($(Q) + (-\lineWidth/4, 0)$);
				\coordinate (O) at ($(N) + (\lineWidth, \lineHeight/2)$);
				\coordinate (R) at ($(O) + (-\lineWidth/2, 0)$);
				\coordinate (S) at ($(R) + (0, \lineHeight)$);
				\coordinate (T) at ($(S) + (-\lineWidth/2, 0)$);			
				\coordinate (U) at ($(T) + (\lineWidth, \lineHeight/2)$);
				\coordinate (V) at ($(U) + (-\lineWidth/2, 0)$);
				\coordinate (W) at ($(V) + (0, \lineHeight)$);
				\coordinate (X) at ($(W) + (\lineWidth*2/3, 0)$);
				\coordinate (Y) at ($(W) + (-\lineWidth/2, 0)$);
				\coordinate (Z) at ($(Y) + (\lineWidth*3/2, \lineHeight/2)$);
				\coordinate (AA) at ($(Z) + (-\lineWidth*3/4, 0)$);
				\coordinate (BB) at ($(AA) + (0, \lineHeight)$);
				
				\draw[->, line width=0.3mm] (A) node[below] {$\mathbf{Q}^{(h)}$}-- (D);
				\draw[->, line width=0.3mm] (B) node[below] {$\mathbf{K}^{(h)}$}-- (E);
				\draw[rounded corners=2mm, fill=orange!30!white, draw=black] (F) rectangle (G) node[midway] {MatMul};
				\draw[->, line width=0.3mm] (H) -- (I);
				\draw[rounded corners=2mm, fill=blue!30!white, draw=black] (J) rectangle (K) node[midway] {Scale};
				\draw[->, line width=0.3mm] (L) -- (M);
				\draw[rounded corners=2mm, fill=green!30!white, draw=black] (N) rectangle (O) node[midway] {Mask};
				\draw[->, line width=0.3mm] (P) node[left] {$\boldsymbol{\Phi}$}-- (Q);
				\draw[->, line width=0.3mm] (R) -- (S);
				\draw[rounded corners=2mm, fill=purple!30!white, draw=black] (T) rectangle (U) node[midway] {SoftMax};
				\draw[->, line width=0.3mm] (V) -- (W);
				\draw[->, line width=0.3mm] (C) node[below] {$\mathbf{V}^{(h)}$}-- (X);
				\draw[rounded corners=2mm, fill=orange!30!white, draw=black] (Y) rectangle (Z) node[midway] {MatMul};
				\draw[->, line width=0.3mm] (AA) -- (BB) node[above] {$\mathbf{Y}^{(h)}$};
				
			\end{tikzpicture}
			\caption{Scaled dot-product self-attention with custom masking.}
			\label{fig:self_attention}
		\end{subfigure}
		\caption{MMHA architecture in the PAPC transformer, processing the input through multiple attention heads combined with masking feature to model inter-user relationships and handle pilot contamination.}
		\label{fig:MMHA_Arch}
	\end{figure*}
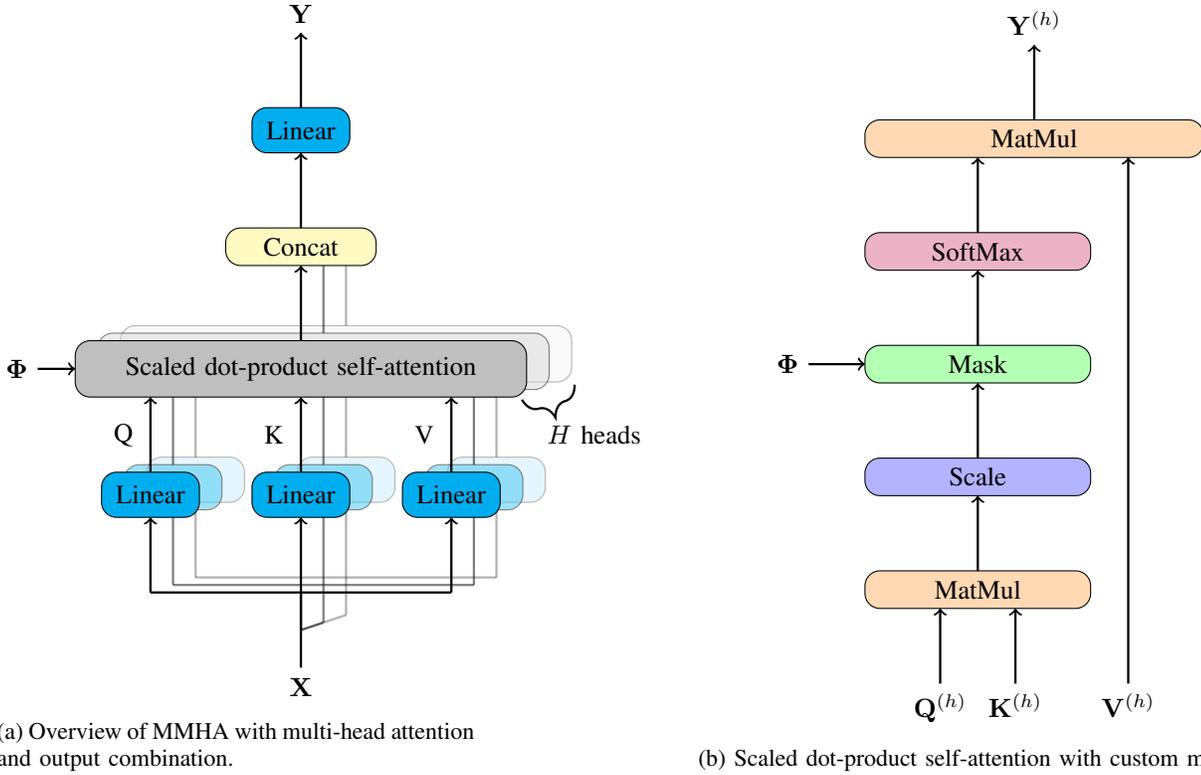
	
	Next, feature-specific scaling and shifting are applied to the normalized matrix $\bar{\mathbf{C}}$ using the trainable vectors $\boldsymbol{\alpha}$ and $\boldsymbol{\beta}$, both of length $\ddot{M}$:
	\begin{align*}
		\text{LayerNorm}(\mathbf{C}; \mathbf{W}_{L}) &= \left(\mathbf{1}_{K} \boldsymbol{\alpha}^{T}\right) \odot \bar{\mathbf{C}} + \mathbf{1}_{K} \boldsymbol{\beta}^{T},
	\end{align*}
	where $\mathbf{W}_{L}=\{\boldsymbol{\alpha}, \boldsymbol{\beta}\}$ represents the trainable parameters.
	
	Unlike in GPT, where each input feature vector is normalized independently, the PAPC transformer normalizes all feature vectors together, preserving inter-user relationships. Additionally, it applies feature-specific scaling and shifting for greater flexibility across feature dimensions, unlike GPT’s scalar-based approach.
	
	\subsubsection{Preprocessing Stage} \label{Preprocessing}
	The preprocessing stage of the PAPC transformer is designed to prepare the input large-scale fading coefficients matrix $\mathbf{B}$ for use within the transformer architecture. The matrix is first transposed so that each row corresponds to a user, allowing the model to treat each user's fading coefficients as a single unit. All the linear layers in the transformer architecture are thus performed row-wise.
	
	To handle the significant variation in the values of $\mathbf{B}$, an element-wise logarithm is applied. Each row of $\mathrm{ln}({\mathbf{B}^{T}})$ is linearly mapped into a higher-dimensional space of length $\bar{M}$ ($\bar{M} > M$), extracting a richer set of features. Layer normalization is applied before and after the mapping to ensure stable training. Let $\mathbf{W}_{P}$ represent all the trainable parameters in this preprocessing step, then the output is denoted as $\mathbf{Z}^{(0)} = \mathbf{F}_{P}(\mathbf{B}; \mathbf{W}_{P})$. Fig~\ref{fig:preprocessing} depicts this functionality.
	
	In NLP models like GPT, preprocessing involves embedding discrete tokens into continuous vectors and applying positional encoding. In the PAPC transformer, each user’s fading coefficients are treated similarly to tokens, with a linear mapping analogous to the embedding step, transforming each user’s vector into a higher-dimensional space. Semantic similarity between tokens can be compared to channel similarities between users. Since the user order is irrelevant, positional encoding is omitted. The resulting matrix, $\mathbf{Z}^{(0)}$, provides a rich representation of the propagation environment, similar to the context in NLP applications.
	
	The primary computational cost of a single forward pass comes from the linear mapping, with complexity $\mathcal{O}(\bar{M}MK)$.
	
	\def \heightOffset{0}
	\def \lineWidth{2}
	\def \lineHeight{0.5}
	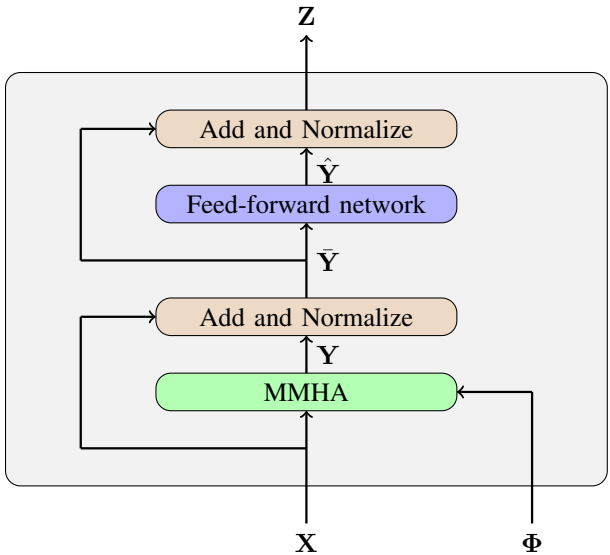
\begin{figure}
		\centering
		\begin{tikzpicture}[scale=1]
			
			\coordinate (A) at (0, 0);
			\coordinate (B) at ($(A) + (0, 2*\lineHeight)$);
			\coordinate (C) at ($(B) + (0, \lineHeight)$);
			\coordinate (D) at ($(B) + (-1.5*\lineWidth, 0)$);
			\coordinate (E) at ($(D) + (0, 3.5*\lineHeight)$);
			\coordinate (F) at ($(E) + (0.5*\lineWidth, 0)$);
			\coordinate (G) at ($(C) + (-\lineWidth, 0)$);
			\coordinate (H) at ($(G) + (2*\lineWidth, \lineHeight)$);
			\coordinate (I) at ($(C) + (0, \lineHeight)$);
			\coordinate (J) at ($(I) + (0, 1*\lineHeight)$);
			\coordinate (K) at ($(J) + (-\lineWidth, 0)$);
			\coordinate (L) at ($(K) + (2*\lineWidth, \lineHeight)$);
			
			\coordinate (M) at ($(A) + (1.5*\lineWidth, 0)$);		
			\coordinate (N) at ($(M) + (0, 3.5*\lineHeight)$);
			\coordinate (O) at ($(N) + (-0.5*\lineWidth, 0)$);
			
			\coordinate (P) at ($(J) + (0, \lineHeight)$);
			\coordinate (Q) at ($(P) + (0, \lineHeight)$);
			\coordinate (R) at ($(Q) + (0, \lineHeight)$);
			\coordinate (S) at ($(Q) + (-1.5*\lineWidth, 0)$);
			\coordinate (T) at ($(S) + (0, 3.5*\lineHeight)$);
			\coordinate (U) at ($(T) + (0.5*\lineWidth, 0)$);
			\coordinate (V) at ($(R) + (-\lineWidth, 0)$);
			\coordinate (W) at ($(V) + (2*\lineWidth, \lineHeight)$);
			\coordinate (X) at ($(R) + (0, \lineHeight)$);
			\coordinate (Y) at ($(X) + (0, 1*\lineHeight)$);
			\coordinate (Z) at ($(Y) + (-\lineWidth, 0)$);
			\coordinate (AA) at ($(Z) + (2*\lineWidth, \lineHeight)$);
			\coordinate (BB) at ($(Y) + (0, \lineHeight)$);
			\coordinate (CC) at ($(BB) + (0, 2*\lineHeight)$);
			\coordinate (DD) at ($(A) + (-2*\lineWidth, \lineHeight)$);
			\coordinate (EE) at ($(DD) + (4*\lineWidth, 11*\lineHeight)$);
			
			\draw[rounded corners=2mm, fill=gray!10!white, draw=black] (DD) rectangle (EE);
			\draw[line width=0.3mm] (A) node[below] {$\mathbf{X}$}-- (B);
			\draw[->, line width=0.3mm] (B) -- (C);
			
			\draw[line width=0.3mm] (B) -- (D);
			\draw[line width=0.3mm] (D) -- (E);
			\draw[->, line width=0.3mm] (E) -- (F);
			\draw[rounded corners=2mm, fill=green!30!white, draw=black] (G) rectangle (H) node[pos=.5,align=center] {MMHA};
			\draw[->, line width=0.3mm] (I) -- (J) node[anchor=north west] {$\mathbf{Y}$};
			\draw[rounded corners=2mm, fill=brown!30!white, draw=black] (K) rectangle (L) node[pos=.5,align=center] {Add and Normalize};
			
			\draw[line width=0.3mm] (M) node[below] {$\boldsymbol{\Phi}$} -- (N);
			\draw[->, line width=0.3mm] (N) -- (O);

			\draw[line width=0.3mm] (P) -- (Q) node[anchor=west] {$\bar{\mathbf{Y}}$};
			\draw[->, line width=0.3mm] (Q) -- (R);
			
			\draw[line width=0.3mm] (Q) -- (S);
			\draw[line width=0.3mm] (S) -- (T);
			\draw[->, line width=0.3mm] (T) -- (U);
			\draw[rounded corners=2mm, fill=blue!30!white, draw=black] (V) rectangle (W) node[pos=.5,align=center] {Feed-forward network};
			\draw[->, line width=0.3mm] (X) -- (Y) node[anchor=north west] {$\hat{\mathbf{Y}}$};
			\draw[rounded corners=2mm, fill=brown!30!white, draw=black] (Z) rectangle (AA) node[pos=.5,align=center] {Add and Normalize};
			\draw[->, line width=0.3mm] (BB) -- (CC) node[above] {$\mathbf{Z}$};
			
		\end{tikzpicture}
		\caption{The PAPC transformer block processes the input using an MMHA and a feed-forward network, with residual connections and layer normalizations.}
		\label{fig:transformer_block}
	\end{figure}
	
	\subsubsection{MMHA}\label{MMHA_module}
	The MMHA module is the core building block of the PAPC transformer. It processes input data across multiple attention heads and applies masking to capture dynamic inter-user relationships.
	
	The MMHA module takes the matrix $\mathbf{X} \in \mathbb{R}^{K \times \bar{M}}$ and the mask matrix $\boldsymbol{\Phi} \in \mathbb{R}^{K \times K}$ as inputs and produces $\mathbf{Y} \in \mathbb{R}^{K \times \bar{M}}$. The module operates with $H$ attention heads, where $\bar{M}$ is an integer multiple of $H$ ($\bar{M} = HD$; $H$ and $D$ are integers). The overall architecture of the MMHA module is illustrated in Fig.~\ref{fig:MMHA_Arch}.
	
	As shown in Fig.~\ref{fig:MMHA}, for each attention head $h$, $h = 1, \cdots, H$ the input matrix $\mathbf{X}$ is transformed into three matrices: Query ($\mathbf{Q}^{(h)}$), Key ($\mathbf{K}^{(h)}$), and Value ($\mathbf{V}^{(h)}$), each of size $K \times D$, through separate linear transformations. Using these three matrices as input to a scaled dot-product self-attention mechanism, the head then computes the output matrix $\mathbf{Y}^{(h)}$.
	
	The scaled dot-product self-attention mechanism, shown in Fig.~\ref{fig:self_attention}, begins with the computation of attention scores:
	\begin{align*}
		\mathbf{S}^{(h)} = \mathbf{Q}^{(h)}(\mathbf{K}^{(h)})^T/\sqrt{D},
	\end{align*}
	followed by a masking operation:
	\begin{align*}
		\bar{\mathbf{S}}^{(h)} = \mathbf{S}^{(h)} \odot \boldsymbol{\Phi}.
	\end{align*}
	The softmax function \cite{goodfellow2016deep} is then applied row-wise to $\bar{\mathbf{S}}^{(h)}$ to generate the attention weights $\mathbf{A}^{(h)}$, and compute
	
	\begin{align*}
		\mathbf{Y}^{(h)} = \mathbf{A}^{(h)}\mathbf{V}^{(h)}.
	\end{align*}
	
	As shown in Fig.~\ref{fig:MMHA_Arch}, the outputs from all $H$ heads are concatenated along the feature dimension and passed through a linear layer to produce the final output $\mathbf{Y}$, compactly represented as $\mathbf{Y} = \mathbf{F}_{\text{M}}(\boldsymbol{\Phi}, \mathbf{X}; \mathbf{W}_{\text{M}})$, where $\mathbf{W}_{\text{M}}$ denotes all the trainable parameters.
	
	In an attention head, each user's output is computed as a weighted combination of the feature vectors (rows) of the Value matrix, with the contribution of each user's input features determined by data-driven attention weights. Rather than manually crafting input features using element-wise ratios of large-scale fading coefficients, as described in Section~\ref{DNN_inputs}, these heads learn to capture the inter-user relationships from the input data in the form of attention weights. These weights are computed independently across multiple heads, allowing the model to capture diverse relationships between users.
	
	While GPT employs unidirectional attention for sequence generation—where each token is influenced only by its preceding tokens—this is achieved through binary masking, setting attention scores to $-\infty$ for future tokens to ensure their attention weights are zero after the softmax. In contrast, the PAPC transformer requires bidirectional attention, similar to bidirectional encoder representations from transformers (BERT) \cite{devlin2019bert}, to compute all rows of $\mathbf{Y}$ simultaneously, thereby capturing inter-user relationships without sequence dependency. Here, masking is not employed for controlling directionality but rather for incorporating the matrix $\boldsymbol{\Phi}$, which reflects pilot allocation information, whose elements range from $0$ to $1$. The MMHA in PAPC manages user relationships and pilot contamination by integrating BERT's bidirectional attention with GPT-inspired masking.
	
	In GPT, the MMHA module sets masked positions of attention scores to $-\infty$, whereas in the MMHA of PAPC, zeros in $\boldsymbol{\Phi}$ lead to zeros in attention scores while resulting in nonzero attention weights. Intuitively, a zero attention weight is desirable for mutually uncontaminated pair of users, as channel similarities among them is irrelevant. This counterintuitive design, while unexpected, has proven effective based on extensive simulations.
	
	The dominant cost during a single forward pass of MMHA arises from computing the Query, Key, and Value matrices, characterized by $\mathcal{O}(\bar{M}^2K)$.
	
	\def \heightOffset{0}
	\def \lineWidth{1.45}
	\def \lineHeight{2.2}
	\def \nnWidth{4}
	\def \circleOffsetX{5}
	\def \circleOffsetY{4.5}
	\def \NumRectangles{3}
	\def \DeltaX{0.1}
	\def \DeltaY{0.1}
	\def\TextOffset{0.5}
	
	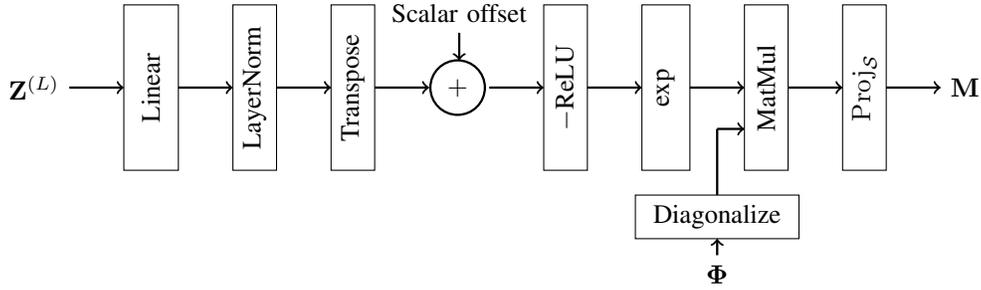
\begin{figure*}
		\centering
		\begin{tikzpicture}[scale=1]
			
			\coordinate (A) at (0, 0);
			\coordinate (B) at ($(A) + (\lineWidth/2, 0)$);
			\coordinate (C) at ($(B) + (0, -\lineHeight/2)$);
			\coordinate (D) at ($(C) + (\lineWidth/2, \lineHeight)$);
			\coordinate (E) at ($(D) + (0, -\lineHeight/2)$);
			\coordinate (F) at ($(E) + (\lineWidth/2, 0)$);
			\coordinate (G) at ($(F) + (0, -\lineHeight/2)$);
			\coordinate (H) at ($(G) + (\lineWidth*0.4, \lineHeight)$);
			\coordinate (I) at ($(H) + (0, -\lineHeight/2)$);
			\coordinate (J) at ($(I) + (\lineWidth/2, 0)$);
			\coordinate (K) at ($(J) + (0, -\lineHeight/2)$);
			\coordinate (L) at ($(K) + (\lineWidth*0.4, \lineHeight)$);
			\coordinate (M) at ($(L) + (0, -\lineHeight/2)$);
			\coordinate (N) at ($(M) + (\lineWidth/2, 0)$);
			\coordinate (P) at ($(N) + (1.1*\lineWidth/4, \lineWidth/4)$);
			\coordinate (O) at ($(P) + (0, \lineHeight/6)$);
			\coordinate (Q) at ($(N) + (1.1*\lineWidth/2, 0)$);
			
			\coordinate (R) at ($(Q) + (\lineWidth/2, 0)$);
			\coordinate (S) at ($(R) + (0, -\lineHeight/2)$);
			\coordinate (T) at ($(S) + (\lineWidth*0.4, \lineHeight)$);
			\coordinate (U) at ($(T) + (0, -\lineHeight/2)$);
			
			\coordinate (V) at ($(U) + (\lineWidth/2, 0)$);
			\coordinate (W) at ($(V) + (0, -\lineHeight/2)$);
			\coordinate (X) at ($(W) + (\lineWidth*0.44, \lineHeight)$);
			\coordinate (Y) at ($(X) + (0, -\lineHeight/2)$);
			
			\coordinate (Z) at ($(Y) + (\lineWidth/2, 0)$);
			\coordinate (HH) at ($(Z) + (0, -\lineHeight/2)$);
			\coordinate (II) at ($(HH) + (\lineWidth*0.4, \lineHeight)$);
			\coordinate (JJ) at ($(II) + (0, -\lineHeight/2)$);
			
			\coordinate (KK) at ($(JJ) + (\lineWidth/2, 0)$);
			\coordinate (LL) at ($(KK) + (0, -\lineHeight/2)$);
			\coordinate (MM) at ($(LL) + (\lineWidth*0.4, \lineHeight)$);
			\coordinate (NN) at ($(MM) + (0, -\lineHeight/2)$);
			
			\coordinate (OO) at ($(NN) + (\lineWidth/2, 0)$);
			
			\coordinate (GG) at ($(Z) + (0, -\lineHeight/4)$);
			\coordinate (FF) at ($(GG) + (-\lineWidth/4, 0)$);
			
			\coordinate (EE) at ($(FF) + (0, -\lineHeight*0.4)$);
			\coordinate (DD) at ($(EE) + (\lineWidth*0.75, 0)$);
			\coordinate (CC) at ($(DD) + (-\lineWidth*1.5, -\lineWidth*0.4)$);
			\coordinate (BB) at ($(CC) + (\lineWidth*0.75, 0)$);
			\coordinate (AA) at ($(BB) + (0, -\lineHeight*0.1)$);
			
			\coordinate (XX) at ($(A) + (\lineWidth/4, -\lineHeight*0.75)$);
			\coordinate (YY) at ($(XX) + (\lineWidth*2.95, \lineHeight*1.5)$);
			
			\draw[->, line width=0.3mm] (A) node[left] {$\mathbf{Z}^{(L)}$}-- (B);
			
			\draw[fill=white] (C) rectangle (D) node[midway, rotate=90] {Linear};

			\draw[->, line width=0.3mm] (E) -- (F);
			\draw[fill=white] (G) rectangle (H) node[midway, rotate=90] {LayerNorm};
			
			\draw[->, line width=0.3mm] (I) -- (J);
			\draw[fill=white] (K) rectangle (L) node[midway, rotate=90] {Transpose};
			
			\draw[->, line width=0.3mm] (M) -- (N);
			
			\pgfmathsetmacro{\halfLineWidth}{\lineWidth/2}
			\node[draw, circle, line width=0.3mm, minimum size=\halfLineWidth cm, right=\lineWidth/2 of M] (add) {$+$};
			
			\draw[->, line width=0.3mm] (O)  node[above] {Scalar offset}-- (P);

			\draw[->, line width=0.3mm] (Q) -- (R);
			\draw[fill=white] (S) rectangle (T) node[midway, rotate=90] {$-$ReLU};
			
			\draw[->, line width=0.3mm] (U) -- (V);
			\draw[fill=white] (W) rectangle (X) node[midway, rotate=90] {exp};
			
			\draw[->, line width=0.3mm] (Y) -- (Z);
			\draw[fill=white] (HH) rectangle (II) node[midway, rotate=90] {MatMul};
			
			\draw[->, line width=0.3mm] (JJ) -- (KK);
			\draw[fill=white] (LL) rectangle (MM) node[midway, rotate=90] {$\mathrm{Proj}_{\mathcal{S}}$};
			
			\draw[->, line width=0.3mm] (NN) -- (OO) node[right] {$\mathbf{M}$};
			
			\draw[fill=white] (CC) rectangle (DD) node[midway] {Diagonalize};
			\draw[line width=0.3mm] (EE) -- (FF);
			\draw[->, line width=0.3mm] (FF) -- (GG);
			\draw[->, line width=0.3mm] (AA) node[below] {$\boldsymbol{\Phi}$} -- (BB);

		\end{tikzpicture}
		\caption{Postprocessing stage of the PAPC, converting the final transformer block's output into power control coefficients through a series of techniques to ensure necessary constraints and to support varying $K$ feature.}
		\label{fig:postprocessing}
	\end{figure*}
	
	\subsubsection{The PAPC Transformer Block}\label{Transformer_block}
	The PAPC transformer block processes the input matrix $\mathbf{X}$ and the mask matrix $\boldsymbol{\Phi}$, producing the output matrix $\mathbf{Z}$ through attention and feed-forward operations, as shown in Fig.~\ref{fig:transformer_block}.
	
	First, the MMHA module computes the intermediate matrix	$\mathbf{Y}$ as $\mathbf{F}_{\text{M}}(\boldsymbol{\Phi}, \mathbf{X}; \mathbf{W}_{\text{M}})$, where $\mathbf{W}_{\text{M}}$ represents its trainable parameters. A residual connection then adds the input $\mathbf{X}$ to $\mathbf{Y}$, and the result is normalized using layer normalization with trainable parameters $\mathbf{W}_{T_1}$, yielding 
	\begin{align*}
		\bar{\mathbf{Y}} = \text{LayerNorm}(\mathbf{X} + \mathbf{Y}; \mathbf{W}_{T_1}).
	\end{align*}
	
	Next, $\bar{\mathbf{Y}}$ is passed through a feed-forward (FF) network with one hidden layer, consisting of $\bar{M}$ units with rectified linear unit (ReLU) activation \cite{goodfellow2016deep}. It computes $\hat{\mathbf{Y}} = \text{FF}(\bar{\mathbf{Y}}; \mathbf{W}_{FF})$ with parameters $\mathbf{W}_{FF}$. Another residual connection adds $\hat{\mathbf{Y}}$ to $\bar{\mathbf{Y}}$, followed by layer normalization with parameters $\mathbf{W}_{T_2}$ to produce
	\begin{align*}
		\mathbf{Z} = \text{LayerNorm}(\bar{\mathbf{Y}} + \hat{\mathbf{Y}}; \mathbf{W}_{T_2}).
	\end{align*}
	Distinct from the GPT architecture, the PAPC transformer block incorporates the external mask matrix $\boldsymbol{\Phi}$ as an input to MMHA.
	
	The overall functionality of the PAPC transformer block is compactly represented as $\mathbf{Z} = \mathbf{F}_{\text{T}}(\boldsymbol{\Phi}, \mathbf{X}; \mathbf{W}_{T})$, where $\mathbf{W}_{T}$ denotes all the trainable parameters. The computational complexity of the PAPC transformer block is dominated by the MMHA and FF modules, with a complexity of $\mathcal{O}(\bar{M}^2K)$ per forward pass.
	
	\subsubsection{Postprocessing Stage}
	The postprocessing stage is designed for the PAPC transformer, consisting of $L$ transformer blocks that output $\mathbf{Z}^{(L)}$ with dimensions $K \times \bar{M}$. The goal is to convert $\mathbf{Z}^{(L)}$ into power control coefficients $\mathbf{M}$, using the additional input $\boldsymbol{\Phi}$, as shown in Fig.~\ref{fig:postprocessing}. 
	
	A linear mapping to $M$ dimensions on each row of $\mathbf{Z}^{(L)}$ is applied:
	\begin{align*}
		\hat{\mathbf{M}} = \text{Linear}(\mathbf{Z}^{(L)}; \mathbf{W}_{O_1}),
	\end{align*}
	where $\mathbf{W}_{O_1}$ represents the trainable parameters. Let $\mathbf{W}_{O_2}$ denote the trainable parameters of a subsequent normalization stage:
	\begin{align*}
		\bar{\mathbf{M}} = \text{LayerNorm}(\hat{\mathbf{M}}; \mathbf{W}_{O_2}).
	\end{align*}
	Next, the matrix $\bar{\mathbf{M}}$ undergoes the following transformation:
	\begin{align*}
		\tilde{\mathbf{M}} = e^{-\text{ReLU}(\bar{\mathbf{M}}^{T} + 6)}
	\end{align*}
   to ensure that each element of $\tilde{\mathbf{M}}$ remains bounded between 0 and 1, and initialized as small positive number. 
	
	The matrix $\tilde{\mathbf{M}}$ is then multiplied by the diagonalized version of $\boldsymbol{\Phi}$, represented as $\tilde{\boldsymbol{\Phi}} = \text{diagonalize}(\boldsymbol{\Phi})$, where the off-diagonal elements of $\boldsymbol{\Phi}$ are set to zero. The final output is obtained by projecting the result onto $\mathcal{S}$:
	\begin{align}
		\mathbf{M} = \mathrm{Proj}_{\mathcal{S}}(\tilde{\mathbf{M}} \tilde{\boldsymbol{\Phi}}). \label{output_projection}
	\end{align}
	The projection operation $\mathrm{Proj}_{\mathcal{S}}(\cdot)$ is performed as a series of per-BS projections, as outlined in \cite{muhammad2021utility}.
	
	Let $\mathbf{W}_{\text{O}}$ represent all the trainable parameters in the postprocessing stage, then, the overall functionality can be represented as $\mathbf{M} = \mathbf{F}_{\text{O}}(\boldsymbol{\Phi}, \mathbf{Z}^{(L)};~\mathbf{W}_{\text{O}})$.
	
	In summary , the linear mapping reduces the dimensionality to match the number of base stations $M$, while the layer normalization ensures stability. The matrix transpose is performed to align with the required $M \times K$ structure of the final output. The ReLU and exponentiation ensure that the elements of $\tilde{\mathbf{M}}$ are in $[ 0, \, 1 ]$, while the scalar shift of $6$ initializes these elements to small positive numbers. The projection operation ensures that the final power control values satisfy the necessary constraints. Finally, the multiplication of $\tilde{\mathbf{M}}$ with the diagonalized $\boldsymbol{\Phi}$ enables the PAPC transformer to flexibly handle a varying number of users avoiding a redesign.
	
	To handle varying user counts (varying $K$ feature), the system supports up to $K_{\text{MAX}}$ users by padding. When $K~<~K_{\text{MAX}}$, $\mathbf{B}$ is padded with a small constant (e.g., $6 \cdot 10^{-13}$) and $\boldsymbol{\Phi}$ with zeros. Combined with the matrix multiplication of $\tilde{\mathbf{M}}$ and diagonalized $\boldsymbol{\Phi}$, this ensures that outputs for users beyond $K$ remain zero, allowing the system, designed for $K_{\text{MAX}}$ users, to adapt without a redesign.
	
	The computational complexity of a forward pass is dominated by the linear mapping layer, which has a complexity of $\mathcal{O}(\bar{M}MK)$.
	
	\subsubsection{PAPC Transformer Model Design}
	\def \heightOffset{0}
	\def \lineWidth{0.5}
	\def \lineHeight{1.2}
	\begin{figure}
		\centering
		\begin{tikzpicture}[scale=1]
			
			\coordinate (A) at (0, 0);
			\coordinate (B) at ($(A) + (2*\lineWidth, 0)$);
			\coordinate (C) at ($(B) + (0, -\lineHeight)$);
			\coordinate (D) at ($(C) + (\lineWidth, 2*\lineHeight)$);
			\coordinate (E) at ($(B) + (\lineWidth, 0)$);
			\coordinate (F) at ($(E) + (\lineWidth, 0)$);
			
			\coordinate (G) at ($(F) + (0, -\lineHeight)$);
			\coordinate (H) at ($(G) + (\lineWidth, 2*\lineHeight)$);
			\coordinate (I) at ($(F) + (\lineWidth, 0)$);
			\coordinate (J) at ($(I) + (3*\lineWidth, 0)$);
			
			\coordinate (K) at ($(J) + (0,-\lineHeight)$);
			\coordinate (L) at ($(K) + (\lineWidth, 2*\lineHeight)$);
			\coordinate (M) at ($(J) + (\lineWidth, 0)$);
			\coordinate (N) at ($(M) + (\lineWidth, 0)$);
			
			\coordinate (O) at ($(N) + (0, -\lineHeight)$);
			\coordinate (P) at ($(O) + (\lineWidth, 2*\lineHeight)$);
			\coordinate (Q) at ($(N) + (\lineWidth, 0)$);
			\coordinate (R) at ($(Q) + (\lineWidth, 0)$);
			
			\coordinate (S) at ($(A) + (0, 1.5*\lineHeight)$);
			\coordinate (T) at ($(S) + (2.5*\lineWidth, 0)$);
			\coordinate (U) at ($(T) + (0, -0.5*\lineHeight)$);
			\coordinate (V) at ($(T) + (2*\lineWidth, 0)$);
			\coordinate (W) at ($(V) + (0, -0.5*\lineHeight)$);
			\coordinate (X) at ($(V) + (4*\lineWidth, 0)$);
			\coordinate (Y) at ($(X) + (0, -0.5*\lineHeight)$);
			\coordinate (Z) at ($(X) + (2*\lineWidth, 0)$);
			\coordinate (AA) at ($(Z) + (0, -0.5*\lineHeight)$);
			
			
			\draw[->, line width=0.3mm] (A) node[left] {$\boldsymbol{\mathbf{B}}$}-- (B);
			\draw (C) rectangle (D) node[pos=.5,align=center,rotate=90] {Preprocessing};
			
			\draw[->, line width=0.3mm] (E) -- (F);
			\draw (G) rectangle (H) node[pos=.5,align=center,rotate=90] {PAPC block $1$};
			
			\draw[->,dashed, line width=0.3mm] (I) -- (J);
			\draw (K) rectangle (L) node[pos=.5,align=center,rotate=90] {PAPC block $L$};
			
			\draw[->, line width=0.3mm] (M) -- (N);
			\draw (O) rectangle (P) node[pos=.5,align=center,rotate=90] {Postprocessing};
			\draw[->, line width=0.3mm] (Q) -- (R) node[right] {$\mathbf{M}$};
			
			\draw[line width=0.3mm] (S)  node[left] {$\boldsymbol{\Phi}$} -- (T);
			\draw[line width=0.3mm]  (T) -- (V);
			\draw[->, line width=0.3mm] (V) -- (W);
			\draw[line width=0.3mm]  (V) -- (X);
			\draw[->, line width=0.3mm] (X) -- (Y);
			\draw[line width=0.3mm]  (X) -- (Z);
			\draw[->, line width=0.3mm] (Z) -- (AA);
		\end{tikzpicture}
		\caption{PAPC transformer architecture using $L$ PAPC transformer blocks. All the blocks and the postprocessing stage incorporate $\boldsymbol{\Phi}$.}
		\label{fig:network}
	\end{figure}
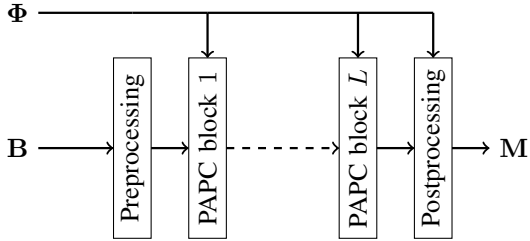
	The PAPC transformer model consists of three main components: the input preprocessing stage, a sequence of $L$ PAPC transformer blocks, and the postprocessing stage. It takes $\mathbf{B}$ as input and produces the power control coefficient matrix $\mathbf{M}$. Fig~\ref{fig:network} represents the PAPC architecture.
	
	As described in Section~\ref{Preprocessing}, the preprocessing stage generates $\mathbf{Z}^{(0)}$ taking $\mathbf{B}$ as input. For $l$ ranging from $1$ to $L$, the output of each transformer block is given by $\mathbf{Z}^{(l)} = \mathbf{F}_{\text{T}}(\boldsymbol{\Phi}, \mathbf{Z}^{(l-1)};~\mathbf{W}^{l}_{\text{T}})$, where $\mathbf{W}^{(l)}_{\text{T}}$ is the trainable parameters of the $l^\text{th}$ block. The postprocessing stage takes $\boldsymbol{\Phi}$ and $\mathbf{Z}^{(L)}$ as input to produce $\mathbf{M}$. Note that all transformer blocks, as well as the postprocessing stage, use the same mask matrix $\boldsymbol{\Phi}$ as an additional input.
	
	Assuming $L \ll M$ and $\bar{M}$ is of order $M$, as observed from the simulations, the computational complexity for a single forward pass in the PAPC transformer is $\mathcal{O}(M^2K)$. In contrast, the computational complexity of the APG is $\mathcal{O}(M_{\text{I}}MK^2)$ \cite{muhammad2021utility}, where $M_{\text{I}}$ is the number of iterations and is comparable to $M$. This comparison highlights a factor of $K$ improvement which underscores a substantial computational advantage of the learning-based PAPC transformer over the APG.
	
	The PAPC transformer is compactly represented as 
	\begin{align*}
		\mathbf{M} = \mathbf{F}_{\text{PAPC}}(\boldsymbol{\Phi}, \mathbf{B};~\mathbf{W}_{\text{PAPC}}),
	\end{align*}
	with $\mathbf{W}_{\text{PAPC}}$ as the trainable parameters. The transformer $\mathbf{F}_{\text{PAPC}}$ is trained to approximate the optimal mapping from the large-scale fading matrix and the pilot allocation information matrix to power control coefficients that maximize the empirical average of smoothed minimum SE as given in \eqref{opt_problem_empirical}. By leveraging attention mechanism to capture structural relationships in $\mathbf{B}$ and incorporating $\boldsymbol{\Phi}$, it offers a scalable and efficient solution for downlink power control in CFmMIMO networks.
	\subsubsection{Training the DNN}
	The PAPC is trained using the PyTorch library, which automatically handles the backpropagation.\footnote{We thank Andrei Palshin, Department of Information and Communications Engineering, Aalto University, for his invaluable support in training, fine-tuning, and validating the deep learning models used in this work.} The adaptive moment estimation (ADAM) optimizer is employed, configured with parameters $\beta^{\text{ADAM}}_{1} = 0.9$, $\beta^{\text{ADAM}}_{2} = 0.98$, and $\epsilon^{\text{ADAM}} = 10^{-9}$, as suggested in \cite{ashish2017attention}. Training is conducted over $16$ epochs with mini-batches of size $1024$. A learning rate scheduler adjusts the learning rate ($l_{\text{rate}}$) based on the training step number $n_{\text{step}}$, as in \cite{ashish2017attention}:
	\begin{align}
		l_{\text{rate}} = d_{\text{mod}}^{-0.5} \cdot \min(n_{\text{step}}^{-0.5}, n_{\text{step}} \cdot n_{\text{warmup}}^{-1.5})
	\end{align}
	where $n_{\text{warmup}}$ is set to $4000$, and $d_{\text{mod}}$ depends on the model's size.\footnote{In Section~\ref{sec:num_results}, $d_{\text{mod}}$ is set according to the simulation scenarios considered. The complete implementation is available in the GitHub repository \cite{k2024cfmmimo}.}
	
	\section{Numerical Results} \label{sec:num_results}
	\subsection{Simulation Setup}
	The performance of the PAPC transformer in downlink power control is evaluated in a CFmMIMO system with a density of $1000$ BSs per sq. km. Various scenarios with different numbers of BSs and users, where each BS is equipped with $N=4$ antennas, is considered. A wrap-around topology is assumed to simulate a large area to avoid boundary effects.
	
	The distance between the $m^{\text{th}}$ BS and the $k^{\text{th}}$ user is $d_{mk}$~[km]. A three-slope path loss model defines the path loss: $PL_{mk} = -L_0 - 15 \log_{10}(d_1) - 20 \log_{10}(d'_{mk})$~[dB], where $d'_{mk}$ is
	\begin{align*}
		d'_{mk} = \begin{cases}
			d_0 &\text{$d_{mk} \le d_0$}\\
			d_{mk} &\text{$d_0 < d_{mk} \le d_1$}\\
			d_1 &\text{$d_{mk} > d_1$}.
		\end{cases}
	\end{align*}
	The large-scale fading coefficient of the corresponding channel is $\beta_{mk} = PL_{mk} + z_{mk}$~[dB], with $z_{mk} \sim \mathcal{N}_\mathcal{C}(0, \sigma^{2}_{\rm sh})$ accounting for shadow fading. The parameters are set as $L_0=140.72$ dB, $d_0=0.01$ km, $d_1=0.05$ km, and $\sigma_{\rm sh}=8$~dB, following \cite{ngo2017cellfree}.
	
	The noise power is $P_n = BW 10^{(N_0 + N_f - 30) / 10}$ [W]. Assuming a noise figure of $N_f=9$~dB, the noise power spectral density $N_0 = -173.98$~dBm/Hz, and a channel bandwidth of $BW=20$~MHz, the transmit SNR for the uplink pilot and downlink data are $\zeta_p = 0.2 / P_n$ and $\zeta_d = 1 / P_n$, respectively.
	
	The coherence block and the pilot sequence lengths are $\tau = 200$~symbols and $\tau_p = 20$~symbols, respectively. The pilot allocation method assigns pilots from $\tau_p$ orthogonal sequences, giving the first $\min(K, \tau_p)$ users unique pilots, and then randomly selecting/reusing pilots for the remaining users if $K > \tau_p$. The smoothing parameter in \eqref{utility_fn} is set as $\lambda = 3$. Table~\ref{table:simulation_setup} summarizes the simulation setup.
	\begin{table}
		\centering
		\begin{tabular}{|c|c|}
			\hline
			BS Density												& $1000$ APs per sq. km.\\
			Length of the coherence block							& $200$~symbols\\
			Length of the pilot sequence							& $20$~symbols\\
			$L_0$													& $140.72$~dB\\
			$d_0$													& $0.01$~km\\
			$d_1$													& $0.05$~km\\
			Standard deviation of shadow fading ($\sigma_{\rm sh}$)	& $8$~dB\\
			Noise power spectral density $N_0$						& $-173.98$~dBm/Hz\\
			BandWidth												& $20$~MHz\\
			Total Noise power at the receiver ($P_n$)				& $-91.97$~dBm\\
			Transmit SNR of uplink pilot ($\zeta_p$)				& $1/P_n$\\
			Transmit SNR of downlink data ($\zeta_d$)				& $0.2/P_n$\\
			Smoothening parameter ($\lambda$)						& $3$\\
			\hline
		\end{tabular}
		\caption{Simulation Setup Parameters}
		\label{table:simulation_setup}
	\end{table}
	
	\subsection{Neural Network Models and Training}
	To demonstrate the potential of PAPC, an FCN model is trained alongside PAPC for various scenarios. In all the scenarios, $2000$ samples are used for evaluation, while they are trained on $P=12,000,000$ samples unless stated otherwise. Training occurs on Graphics processing units (GPUs)\footnote{We acknowledge the computational resources provided by the Aalto Science-IT project.}, and testing is done without GPU assistance to ensure fair comparison of computational times across algorithms.
	
	The FCN model uses a flattened $\mathbf{B}$, that becomes a vector of length $MK$, as its input. This is transformed through a layer normalization stage before passing it through three fully connected linear layers, including an input, a hidden, and an output layer. The input and hidden layers are followed by a corresponding layer normalization module and a ReLU unit. The number of features in the hidden layer is $\hat{M}$. Furthermore, a matrix reshaping is performed, followed by a postprocessing operation similar to PAPC's postprocessing module, but without the matrix multiplication used in PAPC to handle the varying $K$ feature.
	
	\subsection{Scenarios and Evaluation Strategy}
	The performance of PAPC is compared against FCN, a simple equal power allocation (EPA) algorithm, and the APG algorithm. In EPA, each BS assigns equal power to all the users in the downlink signal. The empirical cumulative distribution function (CDF) of the per-user SE is used to represent the performance curve of all algorithms. 
	
	Four distinct scenarios are examined, ranging from Scenario~$0$ to Scenario~$3$. Scenario~$0$ represents a small-scale CFmMIMO network with $M=10$ BSs and $K=4$ users within an area of $0.01$ sq. kms. In contrast, Scenarios~$1$, $2$, and $3$ expand the network to $M=100$ BSs, with user counts of $K=20$, $K=40$, and $K=80$, respectively, across an area of $0.1$ sq. kms. For all the scenarios, the number of transformer blocks is $L=3$ and the number of heads is $H=5$.
	
	The use of PAPC enables the extension of CFmMIMO from the small-scale Scenario~$0$, consistent with similar ranges of network sizes discussed in existing literature, to larger configurations considered in Scenarios~$1$ through $3$. In Scenarios~$2$ and $3$, it is important to note that pilot reuse leads to pilot contamination.
	
	To compare the performance of FCN and PAPC, $\hat{M}$ is set in such a way that the number of trainable parameters in both the networks is approximately the same. Due to heavy computational requirements and poor performance, the evaluation of FCN is omitted in Scenario~$3$. To demonstrate the varyink K feature of PAPC, the model for Scenario~$3$ is trained for the values of $K$ between $K_{\text{MIN}}$ and $K_{\text{MAX}}$. Table~\ref{table:scenario_table} summarizes hyperparameters for each scenario. Note that the ability to simulate Scenario $3$ with up to $80$ users highlights the computational efficiency of the proposed PAPC model, showcasing its capacity to handle large-scale CFmMIMO systems.
	
	\begin{table}
		\centering
		\begin{tabular}{|c|c|c|c|c|c|c|}
			\hline
			Scenario 		& $M$ & $K_{\text{MIN}}$ & $K_{\text{MAX}}$ & $\bar{M}$ & $\hat{M}$ & $d_{\text{mod}}$\\
			\hline 
			$0$             & $10$  & $-$  & $4$  & $80$   & $160$ & $16$\\
			\hline
			$1$             & $100$ & $-$ & $20$ & $500$ & $1000$  & $100$\\
			\hline
			$2$             & $100$ & $-$ & $40$ & $500$ & $571$   & $100$\\
			\hline
			$3$             & $100$ & $40$ & $80$ & $500$ & $-$    & $100$\\
			\hline
		\end{tabular}
		\caption{Hyperparameters for Each Scenario}
		\label{table:scenario_table}
	\end{table}
	
	\begin{figure}[ht]
		\centering
		\includegraphics[width=0.48\textwidth, trim={0cm 0cm 0cm 0cm}]{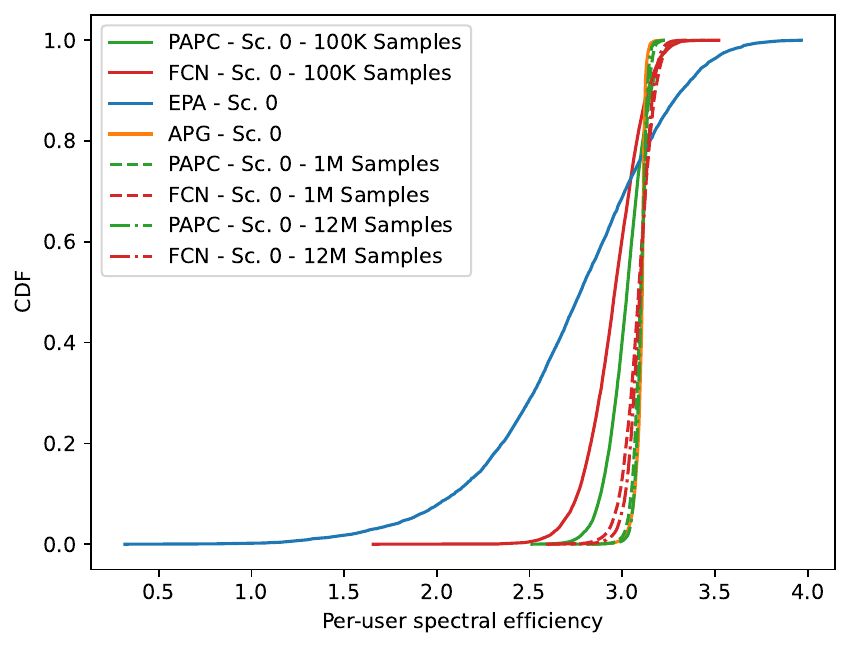}
		\caption{CDF comparison of PAPC, FCN, EPA, and APG in Scenario~$0$ for different training samples. To interpret the results, note that for the max-min fairness objective, a CDF curve that ascends sharply and is also positioned further to the right compared to other curves is considered advantageous. Thus, PAPC outperforms FCN and EPA, approaching APG's performance faster as the number of samples increases.}
		\label{fig:CDFs_Orth}
	\end{figure}
	
	\subsection{Validation of PAPC in Contamination-Free Small-Scale CFmMIMO}
	\begin{figure*}
		\centering
		\includegraphics[width=\textwidth, trim={0cm 0cm 0cm 0cm}]{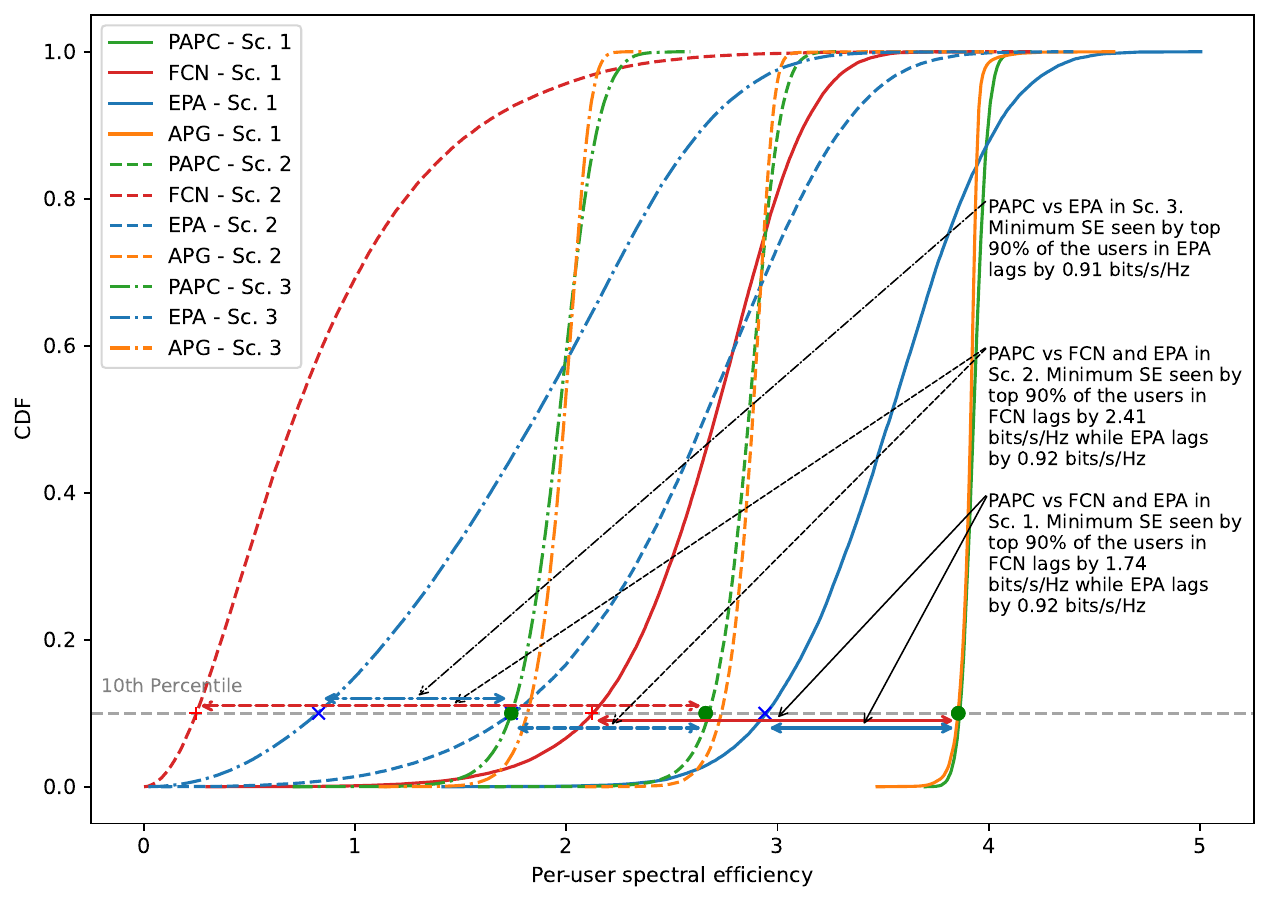}
		\caption{CDF comparison across Scenarios~$1$ to $3$ for different algorithms. PAPC consistently approaches APG performance, outperforming other algorithms due to its masking and attention mechanisms. FCN struggles due to its lack of structure and pilot allocation information.}
		\label{fig:CDFs_LS}
	\end{figure*}
	Scenario~$0$ is a small and contamination-free CFmMIMO system used to assess the PAPC's performance. The FCN and PAPC models are trained with $P=100,000$, $P=1,000,000$, and $P=12,000,000$ samples. For this scenario, Fig.~\ref{fig:CDFs_Orth} compares the CDFs of PAPC with FCN, EPA, and APG.
	
	In this simple and contamination-free scenario, both PAPC and FCN surpass the EPA method with as few as $100,000$ training samples, highlighting the effectiveness of learning-based approaches even with limited data. As the number of training samples increases (from $100,000$ to $1,000,000$ and $12,000,000$), both models continue to improve, with their CDF curves moving closer together, indicating a reduced performance 8gap with the benchmark, APG. PAPC, however, consistently reaches performance closer to APG faster than FCN, likely due to its structural design and effective utilization of pilot information (the fact that there is no contamination). With $12,000,000$ samples, PAPC’s performance closely aligns with that of the APG algorithm, demonstrating its ability to achieve results comparable to the benchmark with sufficient training.
	
	\subsection{PAPC Performance in Large-Scale CFmMIMO Scenarios}
		Scenarios~$1$ to $3$ represent large-scale CFmMIMO systems. Scenario~$1$ is contamination-free, while Scenarios~$2$ and $3$ involve pilot contamination, with Scenario~$3$ experiencing heavy contamination.
		
		The FCN and PAPC models are trained for Scenario~$1$ and Scenario~$2$. Additionally, the PAPC is trained for Scenario~$3$ by enabling the varying $K$ feature. While testing this scenario, the input samples are generated using a fixed number of users, $K=K_{\text{MAX}}$. Fig.~\ref{fig:CDFs_LS} provides a comparative analysis between the CDFs of PAPC, FCN, EPA, and APG.
		
		PAPC consistently approaches the performance of APG in all scenarios, while FCN struggles due to structural inefficiencies, such as the input flattening operation. In Scenario~$2$, pilot contamination further worsens FCN's performance as it lacks pilot allocation information, making it perform worse than both PAPC and EPA. In contrast, PAPC effectively handles pilot contamination, leveraging its masking feature to maintain robust performance.
		
		A quantitative analysis of the minimum SE observed by the top $90\%$ of users further demonstrates PAPC's superiority. In Scenario~$1$, FCN lags behind PAPC by $1.74$ bits/s/Hz, and EPA lags by $0.92$ bits/s/Hz. In Scenario~$2$, FCN falls behind PAPC by $2.41$ bits/s/Hz, while EPA lags by $0.92$ bits/s/Hz. Finally, in Scenario~$3$, where heavy contamination is present, EPA lags behind PAPC by $0.91$ bits/s/Hz. Additionally, the minimum SE observed by PAPC lags behind APG by only $0.08$ bits/s/Hz in both Scenario~$2$ and Scenario~$3$, while for Scenario~$1$, the difference is negligible\footnote{This detail is not annotated in the figure to avoid further visual congestion.}. These results emphasize the robustness of PAPC across different levels of contamination, providing consistent gains compared to other methods.
		
		Scenario~$3$ is the most critical one for evaluating PAPC, as it subjects the model to severe pilot contamination. While being trained with the varying $K$ feature, PAPC shows strong test performance with a fixed number of users, effectively managing the contamination. Since Scenario~$3$ addresses the case of large-scale configuration, this scenario is used to compare the computational efficiency of PAPC, EPA, and APG. Table~\ref{table:Run-time} details the average computational time for each approach, measured on a $64$-bit Windows-$10$ system with $16$~GB RAM and an Intel(R) Xeon(R) Platinum 8176, $2.10$~GHz, without GPU testing.
		
		From the table, the computationally inefficient EPA is the fastest algorithm, while PAPC achieves comparable performance to that of APG, but it is nearly $1000$ times faster than APG.
	\begin{table}
		\centering
		\begin{tabular}{|c|c|}
			\hline
			Algorithm & Run-time (in secs)\\
			\hline 
			APG             & $38.7373$\\
			PAPC             & $0.0262$\\
			EPA             & $0.0003$\\
			\hline
		\end{tabular}
		\caption{Run-time of the algorithms in Scenario~$3$. EPA is the fastest algorithm, while PAPC achieves comparable performance to that of APG, but it is nearly $1000$ times faster than APG.}
		\label{table:Run-time}
	\end{table}
	\begin{figure}[ht]
		\centering
		\includegraphics[width=0.48\textwidth, trim={0cm 0cm 0cm 0cm}]{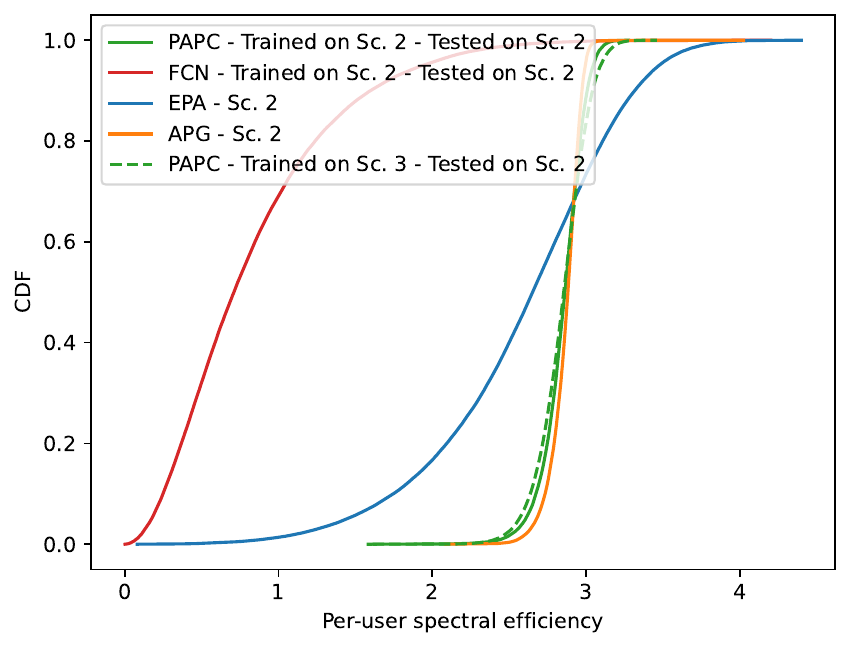}
		\caption{Comparison of PAPCs trained on Scenario~$2$ and Scenario~$3$ (tested with $K=40$), showing matching performance on Scenario~$2$ and validating that larger configurations with padding do not compromise results.}
		\label{fig:CDFs_scenario_comp}
	\end{figure}
 
	\subsection{Evaluating the Flexibility of PAPC}
	Fig.~\ref{fig:CDFs_scenario_comp} compares PAPC trained in Scenario~$3$ with varying K feature enabled, evaluated with fixed $K=40$, to PAPC trained and tested in Scenario~$2$ with fixed $K=40$. The performance of PAPC in both cases is nearly identical, validating that the padding and postprocessing tricks for handling varying K feature do not compromise the performance.
	
\begin{figure}[ht]
		\centering
		\includegraphics[width=0.48\textwidth, trim={0cm 0cm 0cm 0cm}]{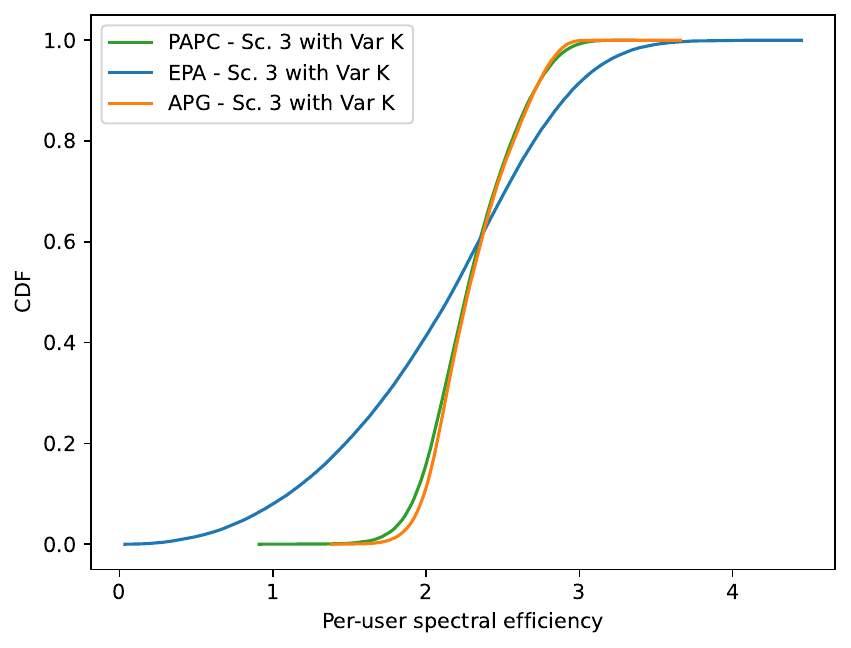}
		\caption{PAPC performance in Scenario~$3$ with varying $K$, comparable to APG and surpassing EPA. PAPC maintains its strong performance, matching the APG algorithm and outperforming EPA, demonstrating its ability to dynamically adjust to fluctuating user counts without any loss in efficiency.}
		\label{fig:CDFs_varK}
	\end{figure}

	Fig.~\ref{fig:CDFs_varK} presents the performance of PAPC when tested in Scenario~$3$ with the varying $K$ feature enabled. The results show that PAPC maintains its strong performance, matching the APG algorithm and outperforming EPA, demonstrating its ability to dynamically adjust to fluctuating user counts without any loss in efficiency.
	
	\section{Conclusion} \label{conclusion}
	The proposed PAPC transformer offers an innovative and efficient solution for downlink power control in CFmMIMO by utilizing the attention mechanism to leverage inter-user relationships and incorporating pilot allocation information via a novel masking technique. This enables PAPC to handle pilot contamination effectively, a limitation that confined prior learning-based methods to small-scale systems. By demonstrating scalability to a system size as large as $MK = 8000$—the first in literature of learning-based solutions—PAPC significantly outperforms FCNs and matches the performance of traditional algorithms like APG with far greater computational efficiency. The PAPC's ability to adapt to varying numbers of users provides additional flexibility. With computational efficiency up to $1000$ times faster than APG, PAPC offers a scalable solution for power control, with the potential to extend beyond the systems explored here, though scalability remains dependent on available training resources.

        \scriptsize
        \small
        \balance
	\bibliographystyle{IEEEtran}
	\bibliography{main}
\end{document}